\documentclass{article}

\usepackage{arxiv}
\date{}
\usepackage[utf8]{inputenc} 
\usepackage[T1]{fontenc}    
\usepackage{url}            
\usepackage{booktabs}       
\usepackage{amsmath}        
\usepackage{amsfonts}       
\usepackage{amssymb}        
\usepackage{nicefrac}       
\usepackage{microtype}      
\usepackage{graphicx}
\usepackage{xcolor}         
\usepackage{multirow}
\usepackage{caption}        
\usepackage{subcaption}
\usepackage{float}
\usepackage{algorithm}
\usepackage{algorithmic}
\usepackage[numbers,sort&compress]{natbib}   
\usepackage[colorlinks=true,allcolors=blue,linktoc=page]{hyperref}   

\title{\textbf{Retraining-Free Depth Pruning of Vision Transformers: A Cross Domain Study of Layer Redundancy}}

\author{
  Alessandro Viespoli\thanks{Corresponding author.} \\
  University of Padua \\
  \texttt{alessandro.viespoli@studenti.unipd.it} \\
  \And
  Loris Nanni \\
  University of Padua \\
  \texttt{loris.nanni@unipd.it} \\
}

\begin{document}
\maketitle
\addtocontents{toc}{\protect\setcounter{tocdepth}{-1}}

\begin{abstract}
Transformer architectures have achieved strong results in natural language processing and computer vision, but their deployment is limited by high computational, memory and storage costs, motivating research into model compression techniques, such as pruning, while preserving performance. Prior work has shown substantial redundancy in large language models (LLMs), and redundancy in vision transformers has also been investigated, but it remains less systematically analyzed than in LLMs. In this work, we extend similarity-based redundancy analysis to vision models by measuring cosine similarity between layer inputs and outputs to identify transformations with minimal impact. We study redundancy patterns across several transformer architectures, including ViT, DINOv2, DINOv3 and SwinV2, and find that deep vision transformers such as DINOv2 and SwinV2, like language models, exhibit significant redundancy. We then evaluate how this redundancy can be exploited to improve efficiency, analyzing the trade-off between inference speed and performance degradation under different pruning strategies. Our results show that moderate pruning can yield substantial acceleration with limited impact on accuracy, and that 20\% of the DINOv2 blocks can be removed without any retraining, for a 1.24$\times$ throughput gain and a 20\% lower energy per image at a relative accuracy loss below 1\% on ImageNet-1K. Across 12 downstream datasets, statistical tests confirm that moderate pruning of DINOv2 and SwinV2 preserves the accuracy of the unpruned model within 2 percentage points, and combining depth pruning with weight quantization reduces computation and memory together, with DINOv2 weights up to 16$\times$ smaller at 4 bits. We further propose a model selection pipeline that automatically identifies the best pruned configuration for a given dataset, reducing FLOPs by 13\% to 27\% on average across architectures with an average accuracy change between $-0.3$ and $+1.0$ points. Our aim is not to rank architectures against one another but to show that pruning yields compact, task-adapted models, supporting efficient deployment and practical use in real-world scenarios.
\end{abstract}


\section{Introduction}
Transformer architectures have become the dominant paradigm in modern artificial
intelligence, achieving state-of-the-art performance in natural language processing
\citep{vaswani2017attention} and computer vision
\citep{dosovitskiy2021imageworth16x16words}. However, these models typically contain
hundreds of millions to billions of parameters and demand substantial memory and
computational power, limiting their deployment in resource-constrained settings such as
edge devices, real-time inference and latency-sensitive applications
\citep{sun2024simpleeffectivepruningapproach, lin2024awqactivationawareweightquantization}.
This limitation is particularly acute for vision transformers, which are increasingly
adopted in domains such as autonomous driving, medical imaging and remote sensing
\citep{zhu2021visiontransformerpruning, yu2022unifiedvisualtransformercompression}.

A growing body of work has addressed model efficiency through pruning \citep{han2015learning, li2017pruningfilters}, quantization
\citep{frantar2023gptqaccurateposttrainingquantization} and knowledge distillation
\citep{hinton2015distillingknowledgeneuralnetwork}. Among these, layer-level structural
pruning is particularly attractive for practical deployment. Unlike quantization or distillation, which require specialized inference kernels or separate training procedures, removing entire layers produces a standard model that is directly compatible with existing deployment pipelines. Recent studies on large language models (LLMs) have shown that deeper layers often perform near-identity transformations and that a substantial fraction of model depth
can be removed with minimal performance loss
\citep{men2024shortgptlayerslargelanguage, gromov2025unreasonableineffectivenessdeeperlayers}.
In vision transformers, training-free depth reduction has so far been studied at the level of whole blocks \citep{cannistraci2026toast}. How much redundancy pretrained vision backbones expose at the finer granularity of individual attention and feed-forward sublayers, across architectures and visual domains, has not yet been systematically characterized.

In this paper, we adapt the LLM-Drop framework
\citep{he2024matterstransformersattentionneeded}, a training-free pruning method
originally designed for LLMs, to vision transformer architectures.
LLM-Drop identifies redundant layers by measuring the cosine similarity between their inputs and outputs, since a layer with high similarity performs a near-identity mapping and can be removed with negligible impact.
The method operates at sublayer granularity, independently
targeting attention or multilayer perceptron (MLP) components. We extend this framework to vision architectures with fundamentally different structures, including encoder-only models and hierarchical models with stage-wise resolution changes
\citep{dosovitskiy2021imageworth16x16words, dinov2, simeoni2025dinov3, swinv2}. We provide a systematic empirical characterization of layer redundancy in vision transformers across architectures and domains, and show that, without retraining, its patterns differ from those reported for LLMs. We also quantify the throughput gains, the energy savings and the reduction in floating-point operations (FLOPs) obtained by layer removal, and show that depth pruning can be combined with post-training weight quantization to further reduce the memory footprint of the model.

Beyond characterizing redundancy, a practical obstacle stands between these findings and their deployment. Even when pruning is effective, identifying the optimal configuration for a specific downstream task requires navigating a combinatorial space of architecture, pruning strategy and number of removed layers. No general solution exists a priori, as the best configuration depends on dataset-specific characteristics \citep{benmeziane2021comprehensivesurveyhardwareawareneural}. There is therefore a practical need for automated and inexpensive methods that turn a pretrained backbone into a compact, task-adapted model. To address this need, we propose an automated model selection pipeline that, given a target dataset, systematically explores the space of pruning configurations and returns a pruned model obtained by training only a linear classification head. The resulting model is a standard transformer that runs on unmodified inference libraries, removing the need for manual experimentation and enabling practitioners to obtain an efficient, task-specific model from a single automated run. We emphasize that the purpose of this study is not to compare architectures or declare a best classifier, but to demonstrate the usefulness of pruning across backbones.

In summary, the key contributions of this paper are as follows:
\begin{itemize}
    \item \textbf{Adaptation of Similarity-Based Pruning to Vision Backbones:} We extend training-free, cosine-similarity layer pruning paradigms, originally conceived for LLMs, to vision transformers. We formalize and evaluate sublayer removal across both flat and hierarchical backbones, including ViT, DINOv2, DINOv3 and SwinV2.
    
    \item \textbf{Comprehensive Cross-Domain Redundancy Benchmark:} Through an empirical evaluation across 13 benchmarks from diverse domains, we show that layer redundancy is a recurring but architecture-dependent property of vision transformers, most pronounced in DINOv2. Notably, when evaluated with its original classification head and no retraining of any kind, DINOv2 tolerates the removal of 20\% of its transformer blocks while yielding a 24\% throughput increase with less than a 1\% relative accuracy reduction.
    
    \item \textbf{Lightweight Adaptation for Performance Recovery:} We show that combining structural layer pruning with lightweight classification head fine-tuning recovers and, in several specialized domains, surpasses unpruned baseline accuracy, while keeping the backbone entirely frozen, reducing FLOPs and increasing throughput.

    \item \textbf{Automated Model Selection Framework:} We propose a task-adaptive, three-phase automated selection pipeline with early stopping. Given a target dataset, the framework explores pruning strategies and drop counts for each architecture, and automatically selects the pruned configuration with the highest validation accuracy under a predefined accuracy tolerance.

    \item \textbf{Observed Divergence from LLM Pruning Behavior:} In our
    experiments without retraining, the redundancy dynamics we observe differ
    from those reported for LLMs. On DINOv2, removing entire
    blocks proves less harmful than removing individual sublayers, the more redundant sublayer type changes from one backbone to another instead of always being attention as reported in LLMs, and Joint Layer Drop mainly benefits the hierarchical SwinV2. These observations suggest that pruning
    guidelines derived from LLMs do not transfer directly to vision
    backbones.

\end{itemize}

The source code is available at \href{https://github.com/zincalex/ViT-Drop}{https://github.com/zincalex/ViT-Drop} or \href{https://github.com/LorisNanni/LLM-Vision-Drop}{https://github.com/LorisNanni/LLM-Vision-Drop}.

The paper is organized as follows: Section~\ref{sec:related_work} reviews related work on model redundancy and compression in language and vision transformers. Section~\ref{sec:materials} presents the evaluated architectures and benchmark datasets. Section~\ref{sec:method} introduces the similarity-based layer dropping framework and pruning strategies. Section~\ref{sec:experiments} details the empirical evaluation, covering
pruning robustness without retraining, fine-tuning performance, statistical
analysis, efficiency trade-offs, the automated model selection pipeline, composability with quantization and a comparison with related methods.
Finally, Section~\ref{sec:conclusion} concludes the paper and outlines future research directions.

\section{Related work}
\label{sec:related_work}
\paragraph{Large Language Models} LLMs have demonstrated strong performance across various tasks. However, the cost of heavy computation and large memory consumption makes their deployment a significant challenge
\citep{sun2024simpleeffectivepruningapproach, lin2024awqactivationawareweightquantization}. The observation that not all
layers in deep transformers contribute equally to the final output has
motivated a growing line of work on structured depth reduction. A first group
of methods identifies removable layers through the similarity between a
layer's input and output, where a value close to 1 indicates a near-identity
transformation. ShortGPT \citep{men2024shortgptlayerslargelanguage}
formalized this intuition as the Block Influence metric, another study located the optimal
contiguous block to remove followed by lightweight fine-tuning \citep{gromov2025unreasonableineffectivenessdeeperlayers} and LLM-Drop
\citep{he2024matterstransformersattentionneeded} refined the granularity to
individual attention and MLP sublayers. A second group replaces similarity
with alternative importance criteria, including per-block entropy
\citep{yang2025entropybasedblockpruningefficient} and gradient magnitudes
from a single backward pass \citep{liu2026gradmapfasterlayerpruning}. A third
group restructures rather than deletes, merging adjacent blocks into a single
wider layer \citep{xu2026flattengptdepthcompressiontransformer} or replacing
removed blocks with a linear mapping estimated from calibration data
\citep{replaceme}. These works collectively establish that structured depth
redundancy is a fundamental property of large transformer architectures,
identifiable through multiple complementary metrics.

\paragraph{Vision Transformers} Vision transformers have achieved competitive
performance across many visual tasks
\citep{touvron2021trainingdataefficientimage, dinov2}, yet their
computational cost limits deployment in resource-constrained settings
\citep{zhu2021visiontransformerpruning, yu2022unifiedvisualtransformercompression}. One family of approaches targets
token-level efficiency, discarding or fusing uninformative tokens through
learned selection modules \citep{rao2021dynamicvitefficient, liang2022evitexpediting} or training-free bipartite matching
\citep{bolya2023tokenmerging}. These strategies reduce per-layer computation
but leave the architecture itself untouched, so every block must still be
traversed and the speedup is bounded by the token reduction ratio.
A second family removes structures from the model itself and recovers accuracy
through gradient-based training. Within it, one line learns importance scores
over projection dimensions, attention heads and blocks, either directly or
through unified optimization and explainability masks
\citep{zhu2021visiontransformerpruning, chen2021chasingsparsityvisiontransformers, yu2022width, yu2022unifiedvisualtransformercompression, yu2023xprunerexplainablepruningvision, 3762387.3762393}; a second line prunes
weight blocks by Taylor saliency followed by full fine-tuning
\citep{xu2025lpvit}; and a third recovers accuracy through distillation or
supervised adaptation on limited data \citep{DcVIT, VBP}. Depth reduction in
particular has recently been shown viable within this family, with
entropy-guided removal of attention sublayers \citep{lambdaVit} and
heterogeneity-aware layer removal \citep{HetDPT}, and SIMPLER \citep{barreiro2026simpler} selects redundant layers from representation similarity on unlabeled task data before fine-tuning. All of these depend on dedicated retraining to restore accuracy.
A third and most recent family operates without any training. CORP \citep{zhang2026corpclosedformoneshotrepresentationpreserving} removes MLP channels and attention dimensions in one shot and compensates for the removed structures through closed-form ridge regression. GRAIL and SNOWS \citep{GRAIL, SNOWS} rely on post-hoc linear reconstruction or Hessian-free second-order preservation, and SnapViT \citep{snapvit} ranks feed-forward neurons and attention heads without labels or a classification head. These methods operate at width granularity, narrowing structures while preserving the depth of the network. TOAST \citep{cannistraci2026toast} instead reduces depth, replacing contiguous runs of whole blocks with the identity or with a closed-form linear map fitted on a few hundred images, and evaluates the result on ViT, DeiT and DINOv2 backbones. Related recovery trade-offs have also been studied for the language backbones of vision-language models \citep{Huang2026StructuralPruningLVLM}.

Across these families, layer removal in vision transformers is either paired with retraining or, in the training-free case of TOAST, restricted to contiguous runs of whole blocks replaced by a fitted mapping. To the best of our knowledge, no study has characterized how much redundancy pretrained vision backbones expose to the removal of individual attention and MLP sublayers without compensation, across flat and hierarchical architectures, pruning strategies and visual domains. Our work addresses this gap by adapting similarity-based sublayer and block pruning to multiple vision architectures and evaluating it on 12 downstream datasets with difference and equivalence tests.

\section{Materials}
\label{sec:materials}
This section presents the vision transformer architectures and datasets used in our
study, detailing the selection criteria and structural characteristics of each model
as well as the composition and splitting protocol of each dataset.

\subsection{Vision Transformer Architectures}
We evaluate the proposed pruning framework on 5 vision transformer architectures, selected to represent a range of design philosophies, model depths and structural organizations. Most selected checkpoints come with a classification head for the 1,000 classes of ImageNet-1K.
\begin{itemize}
    \item ViT-Base \citep{dosovitskiy2021imageworth16x16words} is the first pure transformer architecture to achieve competitive image classification results, and serves as the canonical baseline for subsequent patch-based vision models. It consists of 12 transformer encoder blocks with a hidden
    dimension of 768 and processes $16\times16$ patches from $224\times224$ input images. We
    use the \texttt{google/vit-base-patch16-224} checkpoint, pretrained on ImageNet-21K and
    fine-tuned on ImageNet-1K. Its relatively shallow
    depth makes it a useful reference for evaluating how sensitive compact architectures are
    to layer removal.
    \item DeiT-Base \citep{touvron2021trainingdataefficientimage} shares the ViT-Base architecture, with 12 encoder blocks and a hidden dimension of
    768, but is trained on ImageNet-1K only, using strong data augmentation and regularization in place of the larger pretraining dataset of ViT-Base. We use the \texttt{facebook/deit-base-patch16-224} checkpoint, which is the variant trained without distillation. Since it is architecturally identical to ViT-Base, it provides a controlled contrast for isolating the effect of the training recipe on layer redundancy. 
    \item DINOv2-Giant \citep{dinov2} is a self-supervised vision transformer trained with a
    self-distillation objective on a curated dataset of 142 million images. The model is
    designed as a general-purpose visual feature extractor, producing representations that
    transfer effectively to downstream tasks without requiring
    full fine-tuning of the backbone. We use the
    \texttt{facebook/dinov2-giant-imagenet1k-1-layer} checkpoint, which consists of the
    self-supervised backbone with 40 transformer encoder blocks and a hidden dimension of
    1536, augmented with a single linear classification layer trained on ImageNet-1K. It has the largest depth among the architectures considered in this study, which makes it a natural candidate for studying redundancy in very deep backbones.
    \item DINOv3 ViT-Large \citep{simeoni2025dinov3} builds upon the DINOv2 framework, introducing architectural refinements such as Rotary Position Embeddings (RoPE) for improved spatial reasoning, register tokens to suppress outlier activations, and Gram anchoring to preserve dense feature quality during long training schedules. We use the \texttt{facebook/dinov3-vitl16-pretrain-lvd1689m} checkpoint,
    which consists of 24 encoder blocks with a hidden dimension of 1024. This checkpoint
    provides a pretrained backbone without a classification head. Since DINOv3 is a distilled 
    model derived from a larger teacher within the same
    self-supervised framework, its inclusion allows us to examine how distillation affects
    layer redundancy compared to models trained directly at full scale.
    \item SwinV2-Base \citep{swinv2} is a hierarchical vision transformer that partitions the
    input into non-overlapping windows and applies shifted window attention across multiple
    stages. It introduces residual post-normalization with cosine attention for training
    stability and log-spaced continuous position bias for resolution transfer. We use the
    \texttt{microsoft/swinv2-base-patch4-window16-256} checkpoint, trained on ImageNet-1K at
    $256\times256$ resolution. The model contains 24
    transformer blocks distributed across 4 stages with depths [2, 2, 18, 2] and
    progressively increasing hidden dimensions (128, 256, 512, 1024). Its hierarchical
    structure, with patch merging operations between stages that double the channel dimension
    while halving the spatial resolution, presents a fundamentally different pruning
    challenge compared to flat architectures and tests whether the similarity-based framework
    generalizes beyond uniform-depth models.
\end{itemize}

\subsection{Datasets}
\label{sec:datasets}
We assess the proposed framework by adopting 13 image classification benchmarks from diverse domains. ImageNet-1K serves for the evaluation without retraining of Section~\ref{subsec:zero_shot}, and a subset of Tiny-ImageNet, drawn from the ImageNet-1K training images, serves for calibration, while the remaining 12 datasets are used as downstream benchmarks in the
fine-tuning, efficiency and model selection experiments. Table~\ref{tab:datasets} reports for each dataset the number of classes and samples, the image format, the image size, and the testing protocol. For the
testing protocols, we adopt the following conventions:
\begin{itemize}
    \item OG indicates that the original train/test partition provided by the dataset authors
    is used directly, whether defined via separate files, directory structure, or a predefined permutation matrix, in which case the first permutation is used. The training portion is further split 80:20 stratified per class into training and validation sets.

    \item X:Y:Z indicates the fractions of the total data used for training, validation and testing respectively (e.g., 70:15:15 means 70\% training, 15\% validation and 15\% test). This single-split protocol avoids the computational overhead of full $k$-fold cross-validation, enabling evaluation across a larger number of diverse
    benchmarks within a feasible time budget.
\end{itemize}

The adopted datasets contain information from very different domains, in particular:
\begin{itemize}
    \item ImageNet-1K \citep{5206848} is a large-scale natural image dataset widely used
    for visual recognition, containing 1.28 million training images and 50,000 validation
    images across 1,000 object classes annotated using the WordNet hierarchy. We use the validation set only, employed in its entirety as test data for the experiments of
    Section~\ref{subsec:zero_shot}.
    \item CIFAR-10 \citep{krizhevsky2009learning} consists of 60,000 color images of size
    $32\times32$ pixels divided into 10 object classes (airplanes, automobiles, birds, cats,
    deer, dogs, frogs, horses, ships and trucks).
    \item LCZ42 \citep{1483140} is a remote sensing dataset for Local Climate Zone
    classification, containing 400,673 Sentinel-2 image patches of size $32\times32$ pixels
    from 42 urban agglomerations labeled according to 17 classes. Only the RGB channels are
    used. Training and test splits originate from different geographic regions, creating a
    cross-domain evaluation setting.
    \item CrossD \citep{batrakhanov2024daplanktonbenchmarkdatasetmultiinstrument} is a
    plankton classification dataset containing 44 imbalanced classes captured by different imaging instruments (CytoSense and FlowCytobot). Models are trained on CytoSense images
    and evaluated on FlowCytobot images, constituting a cross-instrument transfer setting.
    \item Zoolake \citep{Kyathanahally_2021} is a dataset of microscopic plankton taxa
    images collected from the dual-magnification Scripps Plankton Camera in Lake Greifensee, built using images acquired from wild plankton between 2018 and 2020. The dataset contains 35 classes.
    \item Laryngeal \citep{sara_moccia_2017_1003200} contains patch images of size
    $100\times100$ pixels extracted from narrow-band imaging endoscopic videos of the
    larynx, divided into 4 classes: IPCL (tissue with intrapapillary capillary loops), Le
    (tissue with leukoplakia), Hbv (tissue with hypertrophic blood vessels) and He (healthy
    tissue).
    \item InfLarynge \citep{Moccia2018LearningBased} consists of 720 manually annotated video frames extracted from 18 narrow-band imaging (NBI) laryngoscopic videos, each acquired from a different patient diagnosed with laryngeal squamous cell carcinoma, with the diagnosis confirmed by histopathological examination. The collected frames are evenly distributed across 4 categories: 180 informative frames, 180 blurred frames, 180 frames affected by saliva or specular reflections, and 180 underexposed frames.
    
    \item Bark \citep{carpentier2018treespeciesidentificationbark} contains
    variable-size RGB images of tree bark textures divided into 23 imbalanced classes.

    \item Pest \citep{deng2018pestrecognition} contains 562 variable-size RGB images of pests commonly found on crops in Europe and Central Asia, divided into 10 balanced classes. The images were collected from several online sources including IPM Images and Mendeley Data.

    \item ColorBG \citep{Dimitropoulos2017GrassmannianVLAD} contains 300 hematoxylin and eosin stained histopathological images collected from patients with invasive ductal carcinoma of the breast. Images were acquired from tumor regions using a microscope with a $40\times$ objective and were annotated according to the Nottingham modification of the Scarff-Bloom-Richardson grading system into Grade 1, Grade 2, and Grade 3.

    \item Kaggle38 \citep{planktonset2015} is a 38-class subset of the PlanktonSet 1.0
    dataset, obtained using the In Situ Ichthyoplankton Imaging System (ISIIS) towed through
    the Straits of Florida. The full dataset was used for the National Data Science Bowl
    2015 competition on Kaggle. Images are grayscale with variable size across 38 imbalanced
    plankton categories.

    \item WHOI22 \citep{orenstein2015whoiplankton} consists of grayscale images of
    size $227\times227$ pixels acquired by the Imaging FlowCytobot. The instrument was deployed in Woods Hole 
    Harbor water,
    producing high-resolution microscope images continuously. The dataset contains 22
    balanced classes.

    \item ZooScan20 \citep{gorsky2010zooscan} is composed of RGB images of variable
    size acquired using the ZooScan integrated system, a flatbed scanner designed for
    digitizing preserved zooplankton samples. Images were collected from a biweekly time
    series in the Bay of Villefranche-sur-mer (France) and automatically segmented into
    individual organisms. The dataset contains 20 imbalanced plankton classes.
\end{itemize}

No data augmentation is applied during training or evaluation. Images are stored and loaded at their native resolution. During preprocessing, the released image processor of each checkpoint resizes them to the input resolution of the corresponding model.

\begin{table}[htbp]
\centering
\small
\caption{Summary of the datasets used in this study.}
\label{tab:datasets}
\begin{tabular}{lccccl}
\toprule
\textbf{Dataset} & \textbf{\#Classes} & \textbf{\#Samples} &
\textbf{Format} & \textbf{Image Size} & \textbf{Protocol} \\
\midrule
ImageNet-1K   & 1000 & 50,000  & RGB       & variable        & - \\
CIFAR-10      & 10   & 60,000  & RGB       & $32\times32$    & OG \\
LCZ42         & 17   & 400,673 & RGB       & $32\times32$    & OG \\
CrossD        & 44   & 94,116  & RGB       & $256\times256$  & OG \\
Zoolake       & 35   & 17,943  & RGB       & variable        & 70:15:15 \\
Laryngeal     & 4    & 1,320   & RGB       & $100\times100$  & OG \\
InfLarynge    & 4    & 720     & RGB       & variable        & OG \\
Bark          & 23   & 23,528  & RGB       & variable        & OG \\
Pest          & 10   & 562     & RGB       & variable        & OG \\
ColorBG       & 3    & 300     & RGB       & $317\times423$  & OG \\
Kaggle38      & 38   & 14,374  & Grayscale & variable        & OG \\
WHOI22        & 22   & 6,598   & Grayscale & $227\times227$  & OG \\
ZooScan20     & 20   & 3,771   & RGB       & variable        & OG \\
\bottomrule
\end{tabular}
\end{table}

\section{Method}
\label{sec:method}
\subsection{Similarity-Based Layer Dropping}
The pruning strategies described in this section were introduced in the LLM-Drop framework \citep{he2024matterstransformersattentionneeded} for structured depth reduction in LLMs. We describe them here because we apply the same strategies to vision transformers.

To quantify the functional contribution of a module, we compare its input $X \in \mathbb{R}^{B \times N \times C}$ with its output $Y \in \mathbb{R}^{B \times N \times C}$, where $B$ is the number of calibration images, $N$ the number of tokens, including the [CLS] token and any register tokens, and $C$ the feature dimension. The cosine similarity is computed for every token and averaged over all images and token positions,
\begin{equation}
    \mathrm{CosineSim}(X, Y) = \frac{1}{B N} \sum_{b=1}^{B} \sum_{i=1}^{N} \frac{\mathbf{x}_{b,i} \cdot \mathbf{y}_{b,i}}{\|\mathbf{x}_{b,i}\|_2 \, \|\mathbf{y}_{b,i}\|_2},
    \label{eq:cosSim}
\end{equation}
so that the score reflects the change of the representation as a whole rather than of individual patches. The importance score of the module is then:
\begin{equation}
    S = 1 - \mathrm{CosineSim}(X, Y).
    \label{eq:importanceScore}
\end{equation}
A similarity close to 1 indicates that the module approximates an identity mapping, which yields a low importance score and marks the module as a candidate for removal.

Formally, let $\mathcal{F}_{\text{Attn}}^l(\cdot)$ and $\mathcal{F}_{\text{MLP}}^l(\cdot)$ denote the attention and MLP functions at layer $l$, operating under Pre-Layer Normalization ($\text{LN}$). When a sublayer is flagged for removal by the pruning mask $\mathbf{m} \in \{0, 1\}$, its transformation reduces to an identity mapping:

\begin{equation}
\begin{aligned}
Y_l = X_l + m_A^l \cdot \mathcal{F}_{\text{Attn}}^l(\text{LN}(X_l)) 
\\
X_{l+1} = Y_l + m_M^l \cdot \mathcal{F}_{\text{MLP}}^l(\text{LN}(Y_l))
\end{aligned}
\end{equation}

where $m_A^l = 0$ (or $m_M^l = 0$) completely bypasses the sublayer without requiring structural modification of adjacent tensor shapes or layer normalization parameters.

In SwinV2, the similarities are computed stage by stage, propagating the calibration activations through each stage and its patch merging layer before the next stage is scored. The scores of all stages are then ranked together, so the removed modules may come from any stage. The patch merging layers, which halve the spatial resolution and double the channel dimension, are never removed.

We evaluate 4 pruning strategies based on which components are scored and removed:
\begin{itemize}
    \item Block Drop measures the similarity between the block input $X_l$ and its final
    output $X_{l+1}$, scoring the entire transformer block as a single unit
    ($S_B^l = 1 - \mathrm{CosineSim}(X_l, X_{l+1})$). Both the attention and MLP sublayers
    are removed simultaneously, providing the most aggressive form of structured pruning.

    \item Attention Drop scores each attention sublayer independently
    ($S_A^l = 1 - \mathrm{CosineSim}(X_l, Y_l)$). Selected attention modules are removed
    while all MLP components are retained.

    \item MLP Drop scores each MLP sublayer independently
    ($S_M^l = 1 - \mathrm{CosineSim}(Y_l, X_{l+1})$). Selected MLP modules are removed
    while all attention components are retained.

    \item Joint Layer Drop collects all sublayer scores into a single global ranking
    $\mathcal{S} = [S_A^1, S_M^1, \ldots, S_A^L, S_M^L]$ and prunes the least important
    modules regardless of type or depth, enabling mixed removal of attention and MLP
    sublayers within a unified importance ordering.
\end{itemize}
Once the importance scores are computed, the candidate modules are ranked and the $n$ least important ones are removed, where $n$ is the target drop count (Algorithm~\ref{alg:similarity_pruning}). Because each module is selected by its own score, the removed modules may lie at arbitrary, non-contiguous depths. LLM-Drop also provides a consecutive variant that removes a single contiguous run of $n$ blocks. We do not use it, because in SwinV2 such a run could span the patch merging layers between stages.

\begin{algorithm}[t]
\caption{Similarity-Based Sublayer Pruning for Vision Transformers}
\label{alg:similarity_pruning}
\begin{algorithmic}[1]
\REQUIRE Pretrained vision transformer model $\mathcal{M}$ with $L$ blocks, Calibration dataset $\mathcal{D}_{\text{calib}}$, Target drop count $n$, Pruning strategy $s \in \{\text{Block}, \text{Attn}, \text{MLP}, \text{Joint}\}$
\ENSURE Pruned model $\mathcal{M}'$

\STATE Initialize candidate score list $\mathcal{S} \leftarrow \emptyset$
\STATE Pass calibration batch $\mathcal{D}_{\text{calib}}$ through $\mathcal{M}$ to extract input/output activations for each layer
\FOR{each transformer block $l = 1, \dots, L$}
    \STATE Extract input $X_l$, post-attention representation $Y_l$, and block output $X_{l+1}$
    \STATE $S_A^l \leftarrow 1 - \text{CosineSim}(X_l, Y_l)$ \COMMENT{Attention sublayer importance}
    \STATE $S_M^l \leftarrow 1 - \text{CosineSim}(Y_l, X_{l+1})$ \COMMENT{MLP sublayer importance}
    \STATE $S_B^l \leftarrow 1 - \text{CosineSim}(X_l, X_{l+1})$ \COMMENT{Full block importance}
\ENDFOR

\IF{$s = \text{Block}$}
    \STATE $\mathcal{S} \leftarrow \{(S_B^1, \text{Block}_1), \dots, (S_B^L, \text{Block}_L)\}$
\ELSIF{$s = \text{Attn}$}
    \STATE $\mathcal{S} \leftarrow \{(S_A^1, \text{Attn}_1), \dots, (S_A^L, \text{Attn}_L)\}$
\ELSIF{$s = \text{MLP}$}
    \STATE $\mathcal{S} \leftarrow \{(S_M^1, \text{MLP}_1), \dots, (S_M^L, \text{MLP}_L)\}$
\ELSIF{$s = \text{Joint}$}
    \STATE $\mathcal{S} \leftarrow \{(S_A^1, \text{Attn}_1), (S_M^1, \text{MLP}_1), \dots, (S_A^L, \text{Attn}_L), (S_M^L, \text{MLP}_L)\}$
\ENDIF

\STATE Sort $\mathcal{S}$ in ascending order by importance score:
 $\mathcal{S}_{\text{sorted}} \leftarrow \text{Sort}(\mathcal{S})$
\STATE Extract the $n$ modules with lowest importance: $\mathcal{P} \leftarrow \mathcal{S}_{\text{sorted}}[1 : n]$

\STATE $\mathcal{M}' \leftarrow \mathcal{M}$
\FOR{each selected module $m \in \mathcal{P}$}
    \STATE Replace module $m$ in $\mathcal{M}'$ with identity mapping function $\mathcal{I}(X) = X$
\ENDFOR

\RETURN Pruned model $\mathcal{M}'$
\end{algorithmic}
\end{algorithm}

\subsection{Pruning Configuration}
Similarity scores are computed offline with a single forward pass over a calibration set $\mathcal{D}_{\text{calib}}$ of $B = 512$ unlabeled images from the training split of Tiny-ImageNet (\cite{le2015tiny}), a 200-class subset of ImageNet-1K downsampled to $64 \times 64$ pixels. $\mathcal{D}_{\text{calib}}$ contains the first 4 images of each of the first 128 classes, and every image is resized by the image processor of each backbone to its input resolution. Since Tiny-ImageNet is drawn from the ImageNet-1K training images, $\mathcal{D}_{\text{calib}}$ does not overlap the ImageNet-1K validation set used for evaluation.
Because similarity estimation requires no gradient computation or backpropagation, this calibration process completes within a few seconds on a single GPU per backbone.

To systematically evaluate trade-offs between depth reduction and downstream task accuracy, we define the search space of each pruning strategy as the set of admissible drop counts $n$. For Block, Attention and MLP Drop the candidate space contains $L$ elements, where $L$ is the number of transformer blocks in the backbone. Joint Layer Drop operates over the union of all attention and MLP sublayers, yielding a space of size $2L$.

The drop counts evaluated differ between the 2 experimental settings. In the analysis without retraining on ImageNet-1K (Section~\ref{subsec:zero_shot}), which involves a single dataset, we sample densely. Every drop count from $n = 1$ to $n = 12$ is evaluated, since this is the range in which vision transformers are most sensitive, followed by larger counts in steps of 4 ($n = 16, 20, \ldots, 32$ and $40$ for DINOv2, and $n = 16, 20, 24$ for SwinV2). ViT-Base and DeiT-Base have 12 blocks and are evaluated at every count, with Joint Layer Drop on DeiT-Base extended to $n = 16$ and $20$. In the downstream benchmark (Sections~\ref{subsec:finetuning} to \ref{subsec:model_selection}), the cross product of 4 architectures, 4 pruning strategies and 12 datasets, each requiring classification head fine-tuning per configuration, makes dense sampling computationally infeasible, so all strategies are evaluated at drop counts in multiples of 4 ($n = 4, 8, \ldots$).

Figure~\ref{fig:vision_similarities} shows the baseline block-level and sublayer-level cosine similarities extracted during calibration, for each architecture as a function of layer index. These structural profiles highlight inherent layer redundancy prior to any fine-tuning or structural adaptation. The exact per-layer values are reported in Appendix~\ref{app:similarities}. 

\begin{center}
\includegraphics[width=\textwidth]{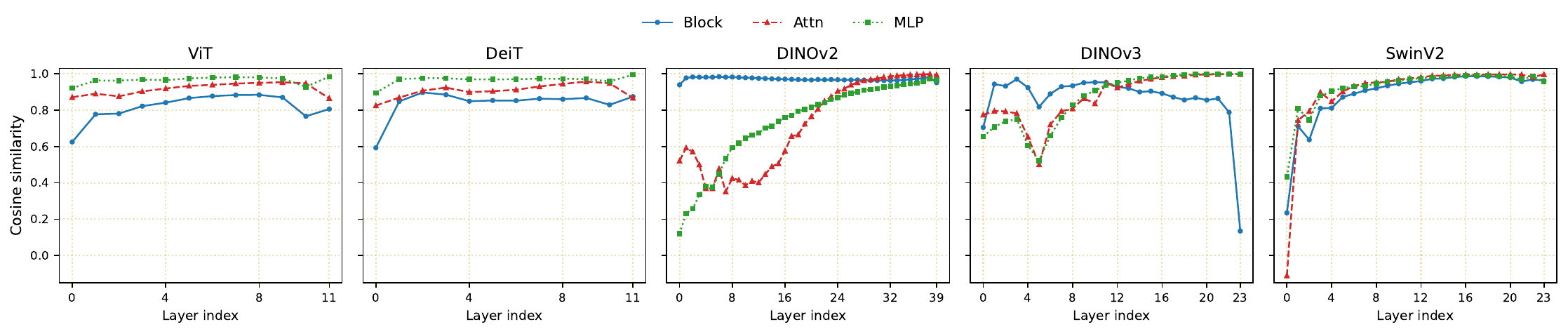}
\captionof{figure}{Per-layer cosine similarities extracted during calibration
for the 5 evaluated backbones. Block denotes the similarity between the full block input and output, while Attn and MLP denote the sublayer-level similarities.}
\label{fig:vision_similarities}
\end{center}

\section{Experimental results}
\label{sec:experiments}
To systematically investigate layer-level redundancy and evaluate the practical efficiency of structured depth reduction in vision transformers, we conduct an empirical study covering 4 representative architectures (ViT, DINOv2, DINOv3 and SwinV2), complemented by DeiT-Base, and 13 visual benchmarks. This section is organized into 7 subsections. First, Section~\ref{subsec:zero_shot} analyzes the intrinsic robustness of
unadapted models to layer removal, isolating architectural redundancy without
any retraining. Second, Section~\ref{subsec:finetuning} examines performance recovery and cross-domain generalization when combining layer pruning with lightweight classification head fine-tuning. Third, Section~\ref{subsec:statistics} tests whether the accuracy differences between pruned and unpruned models are statistically supported. Fourth, Section~\ref{subsec:efficiency} quantifies the computational benefits by measuring inference throughput gains, FLOPs reduction, and the accuracy-speed trade-off via the Speedup Degradation Ratio (SDR) \citep{he2024matterstransformersattentionneeded}. Fifth, Section~\ref{subsec:model_selection} evaluates the proposed automated model selection pipeline in identifying dataset-optimal pruned configurations for practical deployment. Sixth, Section~\ref{subsec:quantization} tests whether depth pruning composes with post-training weight quantization. Finally, Section~\ref{subsec:comparison} positions our results with respect to related structural compression approaches.

All experiments were run on a Slurm-managed GPU cluster equipped with NVIDIA A40
GPUs (48\,GB), inside a Singularity container built on PyTorch~2.7.0 with
CUDA~12.8 and cuDNN~9 (Python~3.11, transformers~5.3.0, accelerate~1.13.0).
Classification-head fine-tuning and all accuracy evaluations were executed in
FP32, data-parallel across 2 GPUs, with fixed seeds for weight initialization
and data shuffling. All efficiency measurements (throughput, latency, memory)
were instead taken on a single dedicated A40 with models cast to BF16; for each architecture, the baseline and all pruned variants were measured on the same
physical device, ensuring that the speedup ratios reported in
Section~\ref{subsec:efficiency} are not affected by hardware variability.

\subsection{Pruning Robustness Without Fine-Tuning}
\label{subsec:zero_shot}
We first evaluated how ViT, DeiT, DINOv2 and SwinV2 respond to pruning without
any retraining. Since these 4 models include a classification head
pretrained on ImageNet-1K, we can directly assess the accuracy of pruned
models on the same dataset without additional adaptation. DeiT-Base is included in this experiment both because it is a widely adopted reference model in the pruning literature and because it shares the ViT-Base architecture while being trained with a different recipe, providing a within-architecture contrast at identical depth. The unpruned baselines match the accuracies reported for
these checkpoints, with DINOv2 at 86.5\%, SwinV2 at 84.7\%, DeiT at 81.8\% and
ViT at 80.3\%. DINOv3 is excluded from this experiment because the checkpoint we adopt does not come with a trained classification head, so evaluating it in this setting would require training one first, which would confound the analysis.

Figure~\ref{fig:VisionDropImagenet1k} reports the accuracy decay as the number
of removed layers increases. ViT degrades rapidly under every strategy, losing 7 points from a single removed block and falling to 3.3\% accuracy after 4 removed blocks and below 25\% after 6 removed sublayers. DeiT-Base, architecturally identical to ViT-Base but trained with a different recipe, is markedly more robust and tolerates roughly twice as many removals before collapse. Its joint ranking removes 11 of its 12 MLP sublayers before selecting any attention module. In contrast to both compact models, DINOv2 and SwinV2 demonstrate considerably greater robustness, consistent with their larger depth.

\begin{center}
\includegraphics[width=\textwidth]{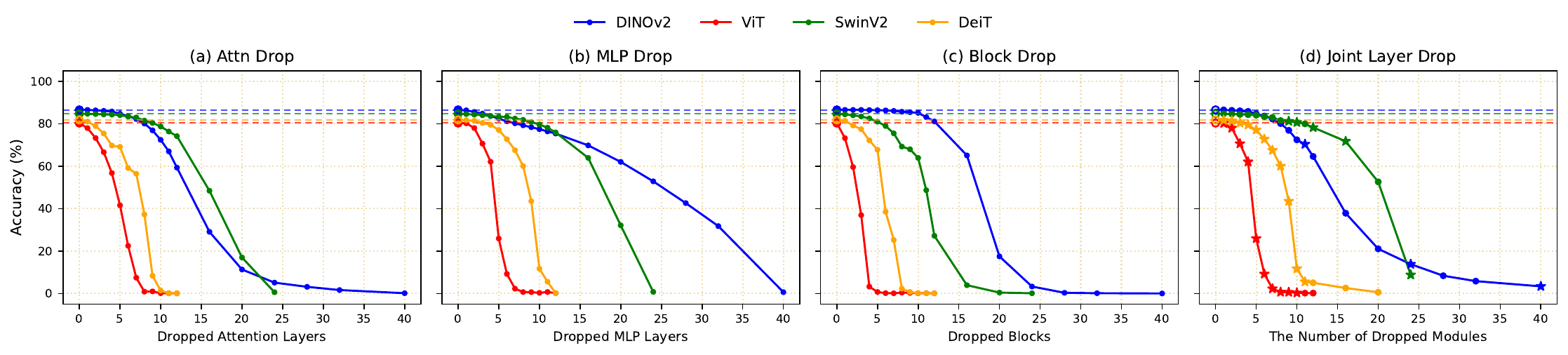}
\captionof{figure}{Accuracy on ImageNet-1K as a function of the number of
removed layers for each pruning strategy, for ViT, DeiT, DINOv2 and SwinV2.
Dashed lines indicate unpruned baseline accuracy. In Joint Layer Drop (d),
\textcolor{red}{$\bigstar$} denotes a removed MLP sublayer and
\textcolor{red}{$\bullet$} denotes a removed attention sublayer.}
\label{fig:VisionDropImagenet1k}
\end{center}
Attention Drop on SwinV2 initially shows the strongest resilience, maintaining higher accuracy than DINOv2 between 7 and 20 dropped
layers. SwinV2 retains 96.2\% of its baseline with 8 attention layers removed,
while DINOv2 retains 97.9\% with 5 removed. Beyond these points, both models deteriorate
rapidly. 

Under MLP Drop, SwinV2 again shows
comparable initial robustness, but DINOv2 degrades more gradually at higher pruning ratios, avoiding the sudden collapse observed in SwinV2.

Block Drop produces unexpected results. DINOv2
retains 98.4\% of its original accuracy after removing 10 entire blocks, whereas removing 10 attention or 10 MLP sublayers leaves 83.7\% and 89.5\% respectively. This is counterintuitive, as removing both sublayers simultaneously discards the most modules per removal step yet yields the least degradation. SwinV2, conversely, is less
robust to block removal, with accuracy declining after only a few blocks are dropped.

SwinV2 benefits most from Joint Layer Drop. Its first 8 removals are attention sublayers, as under Attention Drop, but from then on the joint ranking also selects MLP sublayers, which avoids the collapse of Attention Drop and produces a more gradual degradation. With 16 removed sublayers, SwinV2 still reaches 71.7\% accuracy, 84.7\% of its baseline, before declining more steeply.

Compared to the LLM results reported in the LLM-Drop study \citep{he2024matterstransformersattentionneeded}, on which our pruning framework is based, vision transformers exhibit different redundancy dynamics when evaluated without any fine-tuning. Block Drop is less harmful than removing individual sublayers on DINOv2, whereas LLM-Drop reports it among the most harmful strategies for LLMs. In LLMs, attention sublayers are consistently the most redundant and MLP removal causes severe degradation. In vision transformers, the more redundant sublayer type changes with the backbone, with attention removed first on DINOv2 and SwinV2 and MLP removed first on ViT and DeiT. Finally, Joint Layer Drop mainly benefits the hierarchical SwinV2 rather than being uniformly preferable as in the language setting. These contrasts suggest that pruning guidelines derived from LLMs do not transfer directly to vision architectures.

\subsection{Effect of Fine-Tuning on Pruned Models}
\label{subsec:finetuning}
In practical deployment, vision transformers are rarely used with their original
classification head. Instead, a pretrained backbone is adapted to a target task by
training a new linear classifier while keeping all backbone parameters frozen. This section
investigates whether pruned models can retain their task performance through this
lightweight adaptation.

We evaluate all 4 architectures across the full set of datasets described in
Section~\ref{sec:datasets}. For each pruned variant, the classification head is replaced
with a randomly initialized linear layer matching the number of target classes. Only the classifier is trained using AdamW (learning rate $0.001$, weight decay $0.03$) with a cosine annealing schedule for 5 epochs and batch size 32. The model is then evaluated on
a held-out test set never seen during training. We acknowledge that longer training or full-backbone
fine-tuning could improve absolute accuracy, but our aim is to expose the effect of pruning
rather than to maximize performance. Training only the head for a few epochs keeps the cost low enough to cover the many datasets, architectures, and pruning configurations evaluated
here, under identical conditions for every variant.

Figure~\ref{fig:finetune_retention} summarizes the results of this benchmark. For an architecture $a$ with unpruned baseline accuracy $\mathrm{Acc}_{\text{og}}(a, d)$ on dataset $d$, each pruned configuration with drop count $n$ is scored by its retained accuracy
\begin{equation} R(a, d, n) = 100 \cdot \frac{\mathrm{Acc}_{\text{pruned}}(a, d, n)}{\mathrm{Acc}_{\text{og}}(a, d)} \label{eq:retention} \end{equation}
so a value of 100 means that the pruned model matches its own baseline on that dataset. Each line in Figure~\ref{fig:finetune_retention} reports the median of $R(a, d, n)$ over the 12 datasets, and each shaded band the interquartile range. Curves following the horizontal reference line at 100 therefore indicate pruning without accuracy loss, while portions of a band above it indicate datasets on which the pruned model surpasses its baseline. Overall, fine-tuning the classification head makes all models considerably more robust to pruning than in the unadapted evaluation of Section~\ref{subsec:zero_shot}. In median across datasets, Joint Layer Drop produces the most gradual degradation on every architecture, avoiding the abrupt collapse observed with Block Drop at higher pruning levels. DINOv2 is particularly robust. Under Joint Layer Drop, its median retention is still 98.0\% when 40 of its 80 sublayers are removed. ViT benefits most visibly from fine-tuning, tolerating moderate
pruning levels that were previously catastrophic without any adaptation. DINOv3 stands apart from the other backbones. It is the only one whose median retention stays below 100 at every drop count, and only Joint Layer Drop keeps its median retention above 90\% up to 20 removals. The DINOv3 checkpoint is distilled from a teacher of about 7 billion parameters \citep{simeoni2025dinov3}. We conjecture that compressing such a large teacher into a much smaller student leaves little redundant capacity, so that nearly every layer carries information the network relies on. With a single distilled backbone in our study, this explanation remains a hypothesis.

\begin{center}
\includegraphics[width=\textwidth]{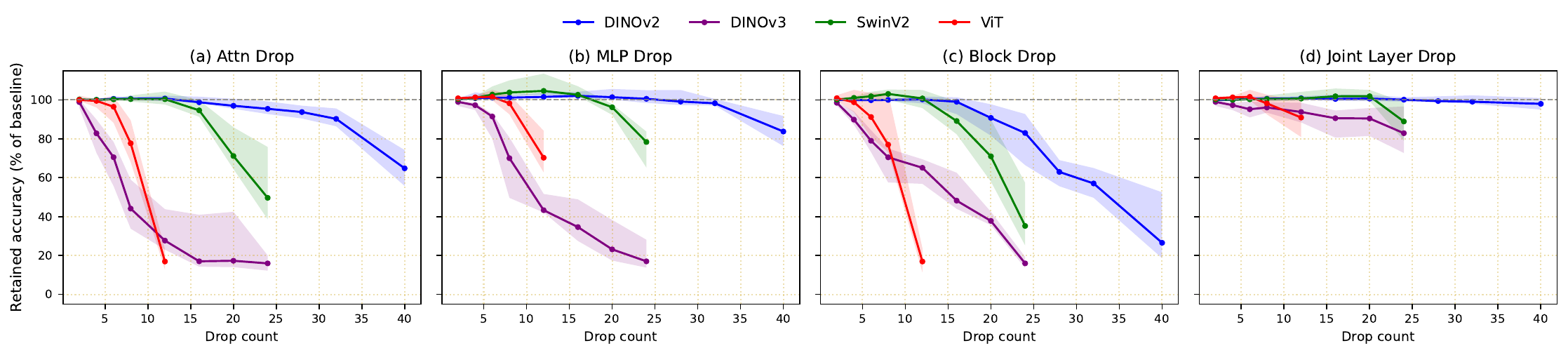}
\captionof{figure}{Retained accuracy under head fine-tuning, aggregated across
the 12 downstream datasets, as a function of drop count for each pruning
strategy. Accuracy of each pruned configuration is normalized by the unpruned
baseline of the same architecture on the same dataset. Solid lines show the median across datasets and shaded bands the interquartile range. The horizontal gray dashed line at 100 marks full baseline retention.}
\label{fig:finetune_retention}
\end{center}

In several cases, pruned fine-tuned models outperform the unpruned baseline. This effect is most frequent on Pest, Laryngeal and InfLarynge for DINOv2 and on Bark, LCZ42, Laryngeal and Pest for SwinV2, where pruned configurations score above the baseline at nearly every drop count from 4 to 12. In Figure~\ref{fig:finetune_retention} it appears as the median retention rising above 100 at moderate drop counts, most clearly under MLP Drop, and also under Block Drop on SwinV2 and under Attention and Joint Layer Drop on DINOv2. On the 2 cross-domain benchmarks, CrossD and LCZ42, pruned configurations also exceed the baseline in most cases on DINOv2 and SwinV2, indicating that in this protocol pruning does not disproportionately harm accuracy under distribution shift.
The complete per-dataset accuracy curves underlying this summary are reported in Appendix~\ref{app:accuracy_curves}.

Beyond accuracy, we record macro-averaged precision, recall and F1-score for every configuration. Two effects dominate these metrics, and both trace back mainly to the deliberately short training budget rather than to pruning itself. Even at drop count 0, the mean gap between accuracy and macro F1-score on the 6 imbalanced benchmarks (Bark, CrossD, Kaggle38, LCZ42, ZooScan20 and Zoolake) is already 9.2 percentage points, since 5 epochs of head training on a skewed class distribution favor the frequent classes. Pruning amplifies this effect only moderately, with the gap growing to 12.3 points at 8 removals and remaining near 12 points thereafter. Precision is at least as high as recall in 81\% of the evaluated configurations, as underfitted heads lose the minority classes first. The clearest illustration of the budget effect is ColorBG, the smallest dataset with only 192 training images. There the DINOv2 baseline head fails to converge within the 5 epochs and collapses to predicting a single class. Its 41.67\% accuracy simply reflects the prevalence of the majority class in the test split, whereas the macro metrics expose the collapse through a recall of exactly one third over 3 classes and a macro F1-score of 0.20. A complete per-dataset view of the 3 metrics, for every architecture and pruning strategy, is provided in Appendix~\ref{app:radar}.

\subsection{Statistical Analysis of Pruned Model Accuracy}
\label{subsec:statistics}
All tests in this section operate on the per-dataset accuracies obtained with the 5-epoch head fine-tuning protocol of Section~\ref{subsec:finetuning}, pairing each pruned configuration with the unpruned baseline of the same architecture on the same dataset. Two complementary tests are used. A difference test establishes whether pruning changes accuracy at all, and an equivalence test establishes whether a pruned model can replace its baseline within a margin that is negligible in practice.

Table~\ref{tab:pvalue_pruning_colonne} reports the p-values of the Wilcoxon signed-rank test \citep{wilcoxon1945individual} comparing each pruned configuration with the unpruned baseline of the same architecture, for all 4 pruning strategies at 4, 8 and 12 removals. The test is performed over the 12 downstream datasets, treating each dataset as a single paired observation. To account for multiple comparisons, the p-values are adjusted with the Holm-Bonferroni correction \citep{holm1979simple}, applied separately to each combination of architecture and pruning method, with the 3 drop counts forming the family of hypotheses.
\begin{table}[t]
  \centering
  \small
  \caption{Holm-adjusted $p$-values of the Wilcoxon signed-rank test between pruning methods and the baseline across 4 architectures. At drop counts 4 and 8, Joint coincides with Attn on DINOv2 and SwinV2 and with MLP on ViT; their adjusted values differ only because the Holm correction also includes drop count 12.}
  \label{tab:pvalue_pruning_colonne}
  \resizebox{\textwidth}{!}{%
  \begin{tabular}{l|ccc|ccc|ccc|ccc}
    \toprule
    & \multicolumn{3}{c|}{\textbf{Joint}} & \multicolumn{3}{c|}{\textbf{Attn}} & \multicolumn{3}{c|}{\textbf{MLP}} & \multicolumn{3}{c}{\textbf{Block}}\\
    \cmidrule(lr){2-4} \cmidrule(lr){5-7} \cmidrule(lr){8-10} \cmidrule(lr){11-13}
    \textbf{Architecture} & \textbf{Drop 4} & \textbf{Drop 8} & \textbf{Drop 12} & \textbf{Drop 4} & \textbf{Drop 8}  & \textbf{Drop 12} & \textbf{Drop 4} & \textbf{Drop 8}  & \textbf{Drop 12} & \textbf{Drop 4} & \textbf{Drop 8}  & \textbf{Drop 12}\\
    \midrule
    DINOv2 & 0.569 & 0.151 & 0.151 & 0.603 & 0.192 & 0.603 & 0.003 & 0.003 & 0.003 & 1.000 & 1.000 & 1.000 \\
    DINOv3 & 0.015 & 0.024 & 0.024 & 0.001 & 0.001 & 0.001 & 0.005 & 0.002 & 0.001 & 0.001 & 0.001 & 0.001 \\
    ViT    & 0.259 & 0.424 & 0.192 & 0.328 & 0.010 & 0.001 & 0.259 & 0.424 & 0.015 & 0.791 & 0.054 & 0.001 \\
    SwinV2 & 0.950 & 0.950 & 0.547 & 1.000 & 1.000 & 1.000 & 0.010 & 0.004 & 0.004 & 0.038 & 0.128 & 0.622 \\
    \bottomrule
  \end{tabular}%
  }
  \vspace{1.5ex}
\end{table}
The p-values separate the architectures cleanly. DINOv2 shows no detectable difference from its baseline under Block, Attention and Joint Layer Drop at any of the 3 drop counts, and SwinV2 behaves the same under Attention and Joint Layer Drop, with Block Drop detectable only at 4 removals. For both backbones, MLP Drop is significant at every drop count, and since the pruned models score above the baseline on most datasets (Section~\ref{subsec:finetuning}), the significance corresponds to an improvement. On ViT, Attention Drop is detectable from 8 removals and Block and MLP Drop only at 12, while Joint Layer Drop shows no detectable difference at any drop count. DINOv3 differs significantly from its baseline in every configuration.

Table~\ref{tab:tost_equivalence} reports the results of the two one-sided tests procedure (TOST) \citep{schuirmann1987tost}, implemented as a pair of one-sided Wilcoxon signed-rank tests, which asks whether the accuracy of the pruned model lies within a fixed margin of the baseline accuracy. The table lists the configurations with a conclusive outcome, and the complete per-configuration statistics, including the inconclusive cases, are reported in Appendix~\ref{app:equivalence}.
We set this margin to 2 percentage points of absolute accuracy before running the analysis. A loss of this size is conventionally regarded as negligible in the compression literature \citep{han2015learning, li2017pruningfilters}, and it is stricter than the tolerance $\tau$ used by our selection pipeline in Section~\ref{subsec:model_selection}. Equivalence is declared at the 0.05 level when both one-sided tests reject, meaning the data support that the true accuracy difference is smaller than 2 percentage points in either direction. To make the analysis independent of this particular choice of margin, the table also reports the Hodges-Lehmann estimate \citep{hodges1963estimates} of the median accuracy difference between the pruned model and the baseline, together with its 90\% confidence interval. The 2 models are equivalent within any margin $\delta$ whenever the entire interval falls between $-\delta$ and $+\delta$, so each reader can apply the margin they consider appropriate.
\begin{table}[htbp]
\centering
\scriptsize
\caption{Conclusive outcomes of the equivalence analysis at drop counts 4, 8 and 12 over the 12 downstream datasets. HL is the Hodges-Lehmann estimate of the median paired accuracy difference (pruned $-$ baseline, percentage points) with its 90\% confidence interval. A configuration is equivalent when the two one-sided Wilcoxon tests against the pre-specified margin of $\pm2$ points both reject at $p < 0.05$. Otherwise, it is superior when its interval lies entirely above zero and the Holm-adjusted two-sided test is significant, and inferior in the symmetric case below zero. Inconclusive configurations are reported in Appendix~\ref{app:equivalence}.}
\label{tab:tost_equivalence}
\begin{tabular}{ll|c|c|c|l}
\toprule
\textbf{Arch} & \textbf{Method} & \textbf{Drop} & \textbf{HL (pp)} & \textbf{90\% CI} & \textbf{Outcome} \\
\midrule
DINOv2 & Block & 4  & $-0.04$ & $[-0.25, +0.17]$ & equivalent \\
DINOv2 & Block & 8  & $+0.01$ & $[-0.94, +0.61]$ & equivalent \\
DINOv2 & Attn  & 4  & $+0.05$ & $[-0.16, +0.82]$ & equivalent \\
DINOv2 & Attn  & 8  & $+0.58$ & $[+0.06, +1.93]$ & equivalent \\
DINOv2 & Joint & 4  & $+0.05$ & $[-0.16, +0.82]$ & equivalent \\
DINOv2 & Joint & 8  & $+0.58$ & $[+0.06, +1.93]$ & equivalent \\
DINOv2 & Joint & 12 & $+0.77$ & $[+0.06, +2.00]$ & equivalent \\
DINOv2 & MLP   & 4--12 & $+1.2$ to $+1.7$ & CIs $>0$ & superior \\
\midrule
SwinV2 & Attn  & 4  & $+0.08$ & $[-0.11, +0.36]$ & equivalent \\
SwinV2 & Attn  & 8  & $+0.44$ & $[-0.70, +1.78]$ & equivalent \\
SwinV2 & Joint & 4  & $+0.08$ & $[-0.11, +0.36]$ & equivalent \\
SwinV2 & Joint & 8  & $+0.44$ & $[-0.70, +1.78]$ & equivalent \\
SwinV2 & MLP   & 4  & $+0.93$ & $[+0.34, +1.85]$ & equivalent \\
SwinV2 & Block & 4  & $+0.83$ & $[+0.44, +2.62]$ & superior \\
SwinV2 & MLP   & 8--12 & $+4.0$ to $+5.4$ & CIs $>0$ & superior \\
\midrule
DINOv3 & all methods & 4--12 & $-2.8$ to $-56.9$ & CIs $<0$ & inferior \\
\midrule
ViT    & Block & 12 & $-66.82$ & $[-75.80, -55.12]$ & inferior \\
ViT    & Attn  & 8--12 & $-17.9$ to $-67.5$ & CIs $<0$ & inferior \\
ViT    & MLP   & 12 & $-23.55$ & $[-31.81, -11.78]$ & inferior \\
\bottomrule
\end{tabular}
\end{table}
The table establishes equivalence for DINOv2 with Block and Attention Drop at 4 and 8 removals and with Joint Layer Drop up to 12 removals, and for SwinV2 with Attention or Joint Layer Drop at 4 and 8 removals and with MLP Drop at 4. In every one of these cases the lower end of the confidence interval lies within 1 percentage point of zero, so the accuracy cost of pruning is bounded well below the margin. MLP Drop fails equivalence in the favorable direction, with confidence intervals entirely above zero on DINOv2 at all 3 drop counts and on SwinV2 at 8 and 12 removals, and SwinV2 Block Drop at 4 removals is superior as well. For DINOv2 MLP Drop, the Hodges-Lehmann estimates place the median gain between 1.2 and 1.7 points, whereas the mean gain over the 12 datasets reaches 4.8 and 5.5 points at 4 and 8 removals, so the mean is driven by a few datasets rather than by a uniform shift. Those datasets are the 2 smallest ones, ColorBG and Pest, on which MLP Drop with 8 removals improves accuracy from 41.67\% to 85.00\% and from 86.05\% to 97.63\% respectively, for the reasons discussed in Section~\ref{subsec:finetuning}. We verified that the statistical superiority of MLP Drop does not hinge on the degenerate ColorBG baseline. Excluding that dataset, the improvement remains significant over the remaining 11 benchmarks at all 3 drop counts (Holm-adjusted $p = 0.006$). DINOv3 is significantly inferior in every configuration, and so is ViT under Attention Drop from 8 removals and under Block and MLP Drop at 12, with intervals entirely below zero.

Reading the 2 tables together adds 3 observations that neither supports alone:
\begin{enumerate}
\item A non-significant Wilcoxon test is not enough to approve a configuration for deployment. DINOv2 and SwinV2 with Block Drop at 12 removals show no significant difference from their baselines in Table~\ref{tab:pvalue_pruning_colonne}, yet they do not reach equivalence, because their confidence intervals, $[-3.58, +1.43]$ and $[-2.11, +3.65]$, extend below the 2 point margin. The data are therefore compatible with no accuracy loss, but also with a loss larger than the tolerance. The same holds for ViT under MLP and Joint Layer Drop at 8 removals, with an interval of $[-6.89, +2.05]$. Conversely, ViT Block Drop at 8 removals has an interval entirely below zero, $[-23.22, -5.29]$, but misses significance after the Holm correction ($p = 0.054$), so it is classified as inconclusive although the data point to a loss. Only the configurations that pass both tests carry a statistically supported bound on the accuracy cost.
\item A high Wilcoxon $p$-value is not a measure of robustness. Block Drop on DINOv2 has $p$-values of 1.000 at every drop count, yet it reaches equivalence only up to 8 removals, while Joint Layer Drop remains equivalent at 12. A large $p$-value only indicates that the data show no consistent direction of change, and at 12 removals Block Drop leaves an interval that straddles zero and extends below the margin.
\item Where the two tests agree, the conclusion is strongest. Attention and Joint Layer Drop at 4 and 8 removals, which select the same modules on both deep backbones, are non-significant under the Wilcoxon test and equivalent under TOST on both DINOv2 and SwinV2.
\end{enumerate}

A final question is whether the similarity ranking itself matters, or whether removing the same number of modules at random would perform equally well. To answer it, we repeat the head fine-tuning protocol of Section~\ref{subsec:finetuning} on 10 configurations of DINOv2, SwinV2 and ViT, replacing the similarity-selected modules with the same number of modules drawn uniformly at random, with 3 random draws per configuration, on 5 datasets spanning the small, medium and cross-domain regimes (Pest, WHOI22, Bark, LCZ42 and CrossD). In this ablation, accuracy is measured on the validation split of each dataset. DINOv3 is excluded since every pruned configuration already degrades significantly. Table~\ref{tab:random_ablation} reports, for each configuration, the mean accuracy change of the similarity-guided and of the random model relative to the unpruned baseline, their difference, and a paired Wilcoxon test over datasets. The per-dataset values are in Appendix~\ref{app:random_ablation}.

Over the 46 configuration and dataset pairs, similarity-guided removal outperforms random removal in 37 cases, by 5.3 points on average and 3.9 in median (Wilcoxon $p < 10^{-6}$). It is also more stable, since the standard deviation across random draws reaches about 8 points on ViT and SwinV2. The advantage, however, is concentrated on the 2 backbones with steep similarity profiles. On ViT and SwinV2, random removal loses 7 to 13 points against the baseline, while similarity-guided removal loses at most 5 points and wins 23 of the 24 pairs, a difference significant on each backbone ($p = 0.002$). With at most 5 datasets per configuration, the smallest attainable two-sided $p$-value of a single configuration is 0.063, so these per-configuration tests cannot reach significance on their own. On DINOv2 the difference is not significant ($p = 0.07$ over 22 pairs, mean gap 2.1 points). This pattern is consistent with the similarity profiles of Figure~\ref{fig:vision_similarities}. Where block similarity varies widely across depth, as in ViT and SwinV2, the ranking separates the near-identity modules from the ones the network relies on, and choosing at random frequently removes one of the latter. Where block similarity is uniformly high, as in DINOv2 with values of at least 0.93 at every layer, almost any subset of modules can be removed at moderate depth, so the ranking has little to discriminate and the number of removed modules matters more than their identity. The ablation therefore supports both parts of our argument. The similarity criterion is what makes depth pruning safe on the compact and hierarchical backbones, and the redundancy of DINOv2 is diffuse rather than concentrated in a few identifiable layers.

\begin{center}
\captionof{table}{Similarity versus random removal of the same number of modules, measured on the validation split over the 5 datasets of the ablation, with 3 random draws per configuration. Similarity and Random are the mean accuracy changes relative to the unpruned baseline in percentage points, Gain is their difference, and $p$ is the paired two-sided Wilcoxon test over datasets. DINOv2 Block-16 was reached on 2 datasets and SwinV2 Joint-8 on 4, since the search phase stopped early on the others.}
\label{tab:random_ablation}
{\small
\begin{tabular}{l l ccc c}
\toprule
 & & \multicolumn{3}{c}{\textbf{Accuracy change (pp)}} & \\
\cmidrule(lr){3-5}
\textbf{Model} & \textbf{Configuration} & \textbf{Similarity} & \textbf{Random} & \textbf{Gain} & \textbf{$p$} \\
\midrule
\multirow{5}{*}{DINOv2}
 & Block-8  & $+0.49$ & $-1.36$  & $+1.85$ & 0.625 \\
 & Block-16 & $-3.36$ & $-0.83$  & $-2.53$ & 0.500 \\
 & MLP-8    & $+0.66$ & $-2.01$  & $+2.67$ & 0.312 \\
 & Joint-8  & $-0.00$ & $-2.47$  & $+2.47$ & 0.188 \\
 & Joint-16 & $+0.98$ & $-2.23$  & $+3.20$ & 0.438 \\
\midrule
\multirow{3}{*}{SwinV2}
 & Block-8  & $-4.94$ & $-10.61$ & $+5.67$ & 0.188 \\
 & MLP-8    & $+1.32$ & $-6.91$  & $+8.23$ & 0.062 \\
 & Joint-8  & $-0.15$ & $-9.05$  & $+8.90$ & 0.125 \\
\midrule
\multirow{2}{*}{ViT}
 & Block-4  & $-4.20$ & $-13.30$ & $+9.10$ & 0.062 \\
 & MLP-4    & $-0.06$ & $-9.18$  & $+9.12$ & 0.062 \\
\bottomrule
\end{tabular}}
\end{center}

\subsection{Efficiency Analysis of Pruned Models}
\label{subsec:efficiency}
We also evaluated how much inference speed can improve when layers are removed, and how
to quantify the trade-off between faster inference and performance degradation. Throughput (images/s) is measured on randomly generated inputs, since it depends only on the model and not on the image content, with a batch size of $B = 128$. After a warm-up phase, each measurement consists of 5 independent runs of $N = 100$ forward passes, and the reported throughput is the mean over the 5 runs, each computed as
\begin{align}
    \mathrm{TH} &= \frac{B}{\frac{1}{N}\sum_{i=1}^{N} t_i} \\[6pt]
    \mathrm{Speedup} &= \frac{\mathrm{TH}_{\text{pruned}}}{\mathrm{TH}_{\text{og}}}
\end{align}
where $t_i$ is the execution time of the $i$-th forward pass of a run. $\mathrm{TH}_{\text{og}}$ and $\mathrm{TH}_{\text{pruned}}$
denote the throughput of the original and pruned models, respectively.
To evaluate the trade-off between accuracy loss and inference acceleration, we use the SDR $\gamma$ of LLM-Drop, computed on the accuracy averaged over the 12 downstream datasets,
\begin{align}
    \Delta_\text{Acc} &= \frac{\mathrm{Acc}_{\text{og}} - \mathrm{Acc}_{\text{pruned}}}{\mathrm{Acc}_{\text{og}}}
    \times 100 \\[6pt]
    \Delta_\text{Speed} &= \frac{\mathrm{TH}_{\text{pruned}} - \mathrm{TH}_{\text{og}}}{\mathrm{TH}_{\text{og}}}
    \times 100 \\[6pt]
    \gamma &= \frac{\Delta_\text{Acc}}{\Delta_\text{Speed}}
\end{align}
where $\mathrm{Acc}_{\text{og}}$ and $\mathrm{Acc}_{\text{pruned}}$ are the average accuracies of the original and pruned models over the 12 datasets.
$\Delta_{\text{Acc}}$ represents the relative accuracy degradation, while
$\Delta_{\text{Speed}}$ represents the relative throughput gain. The quantity $\gamma$
therefore measures the percentage of accuracy lost per 1\% of speedup gained. A value of
$\gamma = 0$ indicates free compression with no accuracy cost, while increasing positive
values indicate progressively less favorable trade-offs. A negative $\gamma$ occurs when the pruned model achieves higher average
accuracy than the unpruned baseline, indicating a Pareto improvement, with
both higher accuracy and higher throughput than the baseline. Because the average includes ColorBG, whose DINOv2 baseline head collapses to a single class (Section~\ref{subsec:finetuning}), the negative $\gamma$ values of DINOv2 largely reflect that single dataset and should be read together with the per-dataset tests of Section~\ref{subsec:statistics}.

We also compute the theoretical number of FLOPs to estimate the reduction in computational cost introduced by pruning. FLOPs are calculated at the input resolution used for every accuracy result, $256 \times 256$ for SwinV2 and $224 \times 224$ for the other backbones. Since the backbones differ in width, depth and patch size, absolute FLOPs should not be compared across architectures, but the relative reductions within each architecture remain meaningful.

\begin{table}[t]
\centering
\caption{Efficiency of pruned models across the 4 architectures. Avg is the mean accuracy over the 12 downstream datasets, Speedup is the throughput relative to the unpruned baseline, $\gamma$ is the SDR, and Energy is the energy per image in joules, computed as the average board power during the throughput runs divided by the throughput. The unpruned baselines reach an average accuracy of 81.99\% for DINOv2, 84.92\% for DINOv3, 80.02\% for SwinV2 and 79.31\% for ViT, with 599.57, 123.11, 41.66 and 35.43 GFLOPs and 2.201, 0.658, 0.693 and 0.155 J per image respectively. Configurations that coincide by construction (Joint with Attn or MLP at 4 and 8 removals) were benchmarked separately, and their speedup differences of about 1\% reflect measurement noise. Dashes mark drop counts not available for a backbone.}
\label{tab:speedup_all}
\resizebox{\textwidth}{!}{%
\begin{tabular}{ll|ccccc|ccccc|ccccc|ccccc}
\toprule
& & \multicolumn{5}{c|}{\textbf{DINOv2}} & \multicolumn{5}{c|}{\textbf{DINOv3}} & \multicolumn{5}{c|}{\textbf{SwinV2}} & \multicolumn{5}{c}{\textbf{ViT}} \\
\cmidrule(lr){3-7} \cmidrule(lr){8-12} \cmidrule(lr){13-17} \cmidrule(lr){18-22}
\textbf{Method} & \textbf{Drop} & Avg & Speedup ($\uparrow$) & $\gamma$ ($\downarrow$) & GFLOPs & Energy & Avg & Speedup ($\uparrow$) & $\gamma$ ($\downarrow$) & GFLOPs & Energy & Avg & Speedup ($\uparrow$) & $\gamma$ ($\downarrow$) & GFLOPs & Energy & Avg & Speedup ($\uparrow$) & $\gamma$ ($\downarrow$) & GFLOPs & Energy \\
\midrule
\multirow{6}{*}{Block}
 & 4  & 82.10 & 1.10$\times$ & $-0.013$ & 540 & 1.993 & 77.18 & 1.19$\times$ & 0.473 & 103 & 0.552 & 81.41 & 1.14$\times$ & $-0.123$ & 35 & 0.610 & 77.29 & 1.43$\times$ & 0.059 & 24 & 0.107 \\
 & 8  & 81.98 & 1.24$\times$ & 0.001 & 480 & 1.772 & 57.07 & 1.48$\times$ & 0.679 & 82 & 0.443 & 82.13 & 1.33$\times$ & $-0.080$ & 28 & 0.523 & 65.58 & 2.62$\times$ & 0.107 & 12 & 0.058 \\
 & 12 & 81.20 & 1.41$\times$ & 0.023 & 420 & 1.556 & 53.24 & 1.96$\times$ & 0.389 & 62 & 0.335 & 80.67 & 1.50$\times$ & $-0.016$ & 22 & 0.463 & 14.29 & 14.71$\times$ & 0.060 & 0.5 & 0.008 \\
 & 16 & 81.01 & 1.64$\times$ & 0.019 & 360 & 1.338 & 46.48 & 2.88$\times$ & 0.241 & 41 & 0.227 & 71.52 & 1.85$\times$ & 0.125 & 15 & 0.376 & -- & -- & -- & -- & -- \\
 & 28 & 51.94 & 3.22$\times$ & 0.165 & 181 & 0.684 & -- & -- & -- & -- & -- & -- & -- & -- & -- & -- & -- & -- & -- & -- & -- \\
 & 40 & 27.70 & 73.85$\times$ & 0.009 & 1.3 & 0.025 & -- & -- & -- & -- & -- & -- & -- & -- & -- & -- & -- & -- & -- & -- & -- \\
\midrule
\multirow{6}{*}{Attn}
 & 4  & 82.37 & 1.03$\times$ & $-0.141$ & 579 & 2.130 & 67.59 & 1.09$\times$ & 2.195 & 116 & 0.602 & 80.37 & 1.08$\times$ & $-0.052$ & 40 & 0.643 & 78.09 & 1.13$\times$ & 0.119 & 31 & 0.137 \\
 & 8  & 83.68 & 1.07$\times$ & $-0.286$ & 558 & 2.052 & 41.42 & 1.21$\times$ & 2.464 & 109 & 0.545 & 80.66 & 1.21$\times$ & $-0.038$ & 37 & 0.576 & 60.39 & 1.30$\times$ & 0.785 & 27 & 0.118 \\
 & 12 & 83.27 & 1.12$\times$ & $-0.134$ & 536 & 1.972 & 30.80 & 1.35$\times$ & 1.802 & 101 & 0.486 & 80.75 & 1.34$\times$ & $-0.027$ & 35 & 0.520 & 15.11 & 1.56$\times$ & 1.447 & 23 & 0.098 \\
 & 16 & 82.59 & 1.16$\times$ & $-0.045$ & 515 & 1.894 & 22.94 & 1.54$\times$ & 1.360 & 94 & 0.428 & 76.42 & 1.54$\times$ & 0.084 & 32 & 0.453 & -- & -- & -- & -- & -- \\
 & 28 & 78.17 & 1.33$\times$ & 0.141 & 452 & 1.653 & -- & -- & -- & -- & -- & -- & -- & -- & -- & -- & -- & -- & -- & -- & -- \\
 & 40 & 55.01 & 1.56$\times$ & 0.592 & 389 & 1.413 & -- & -- & -- & -- & -- & -- & -- & -- & -- & -- & -- & -- & -- & -- & -- \\
\midrule
\multirow{6}{*}{MLP}
 & 4  & 86.83 & 1.06$\times$ & $-0.944$ & 561 & 2.071 & 80.81 & 1.08$\times$ & 0.609 & 110 & 0.609 & 81.21 & 1.03$\times$ & $-0.564$ & 37 & 0.679 & 82.02 & 1.21$\times$ & $-0.160$ & 28 & 0.127 \\
 & 8  & 87.48 & 1.14$\times$ & $-0.484$ & 522 & 1.933 & 55.80 & 1.18$\times$ & 1.949 & 97 & 0.559 & 84.31 & 1.05$\times$ & $-1.061$ & 33 & 0.664 & 76.13 & 1.59$\times$ & 0.068 & 21 & 0.097 \\
 & 12 & 84.36 & 1.23$\times$ & $-0.127$ & 483 & 1.793 & 39.50 & 1.29$\times$ & 1.835 & 83 & 0.509 & 85.13 & 1.08$\times$ & $-0.771$ & 29 & 0.644 & 57.31 & 2.27$\times$ & 0.218 & 13 & 0.067 \\
 & 16 & 85.16 & 1.33$\times$ & $-0.117$ & 444 & 1.655 & 31.95 & 1.43$\times$ & 1.444 & 70 & 0.459 & 83.09 & 1.12$\times$ & $-0.322$ & 24 & 0.624 & -- & -- & -- & -- & -- \\
 & 28 & 84.69 & 1.77$\times$ & $-0.042$ & 328 & 1.240 & -- & -- & -- & -- & -- & -- & -- & -- & -- & -- & -- & -- & -- & -- & -- \\
 & 40 & 68.48 & 2.67$\times$ & 0.099 & 212 & 0.823 & -- & -- & -- & -- & -- & -- & -- & -- & -- & -- & -- & -- & -- & -- & -- \\
\midrule
\multirow{6}{*}{Joint}
 & 4  & 82.37 & 1.03$\times$ & $-0.139$ & 579 & 2.130 & 80.81 & 1.08$\times$ & 0.607 & 110 & 0.609 & 80.37 & 1.08$\times$ & $-0.054$ & 40 & 0.645 & 82.02 & 1.21$\times$ & $-0.161$ & 28 & 0.127 \\
 & 8  & 83.68 & 1.07$\times$ & $-0.284$ & 558 & 2.052 & 80.96 & 1.19$\times$ & 0.247 & 101 & 0.553 & 80.66 & 1.21$\times$ & $-0.038$ & 37 & 0.578 & 76.13 & 1.59$\times$ & 0.068 & 21 & 0.097 \\
 & 12 & 83.18 & 1.12$\times$ & $-0.117$ & 532 & 1.957 & 76.62 & 1.32$\times$ & 0.310 & 91 & 0.500 & 81.54 & 1.28$\times$ & $-0.068$ & 33 & 0.544 & 71.07 & 2.11$\times$ & 0.093 & 15 & 0.072 \\
 & 16 & 84.26 & 1.19$\times$ & $-0.144$ & 502 & 1.847 & 72.17 & 1.48$\times$ & 0.313 & 82 & 0.445 & 82.93 & 1.36$\times$ & $-0.102$ & 29 & 0.512 & -- & -- & -- & -- & -- \\
 & 28 & 82.24 & 1.49$\times$ & $-0.006$ & 399 & 1.476 & -- & -- & -- & -- & -- & -- & -- & -- & -- & -- & -- & -- & -- & -- & -- \\
 & 40 & 83.23 & 1.99$\times$ & $-0.015$ & 296 & 1.105 & -- & -- & -- & -- & -- & -- & -- & -- & -- & -- & -- & -- & -- & -- & -- \\
\bottomrule
\end{tabular}%
}
\end{table}

Table~\ref{tab:speedup_all} reports these metrics for the 4 architectures across
different pruning strategies and numbers of removed layers. Across all architectures, Block Drop produces the highest speedups and the largest FLOPs reduction, since each removal eliminates both sublayers of a block. Attention Drop and MLP Drop remove a single sublayer per step and yield smaller gains, whose ordering depends on the architecture. MLP Drop is faster on DINOv2 and ViT, in line with the larger FLOPs of an MLP sublayer, whereas Attention Drop is faster on SwinV2 and DINOv3, plausibly because their attention involves many small operations, windowing in SwinV2 and RoPE in DINOv3, that cost more wall-clock time than their FLOPs suggest. Negative $\gamma$ values, which indicate a pruned model more accurate on average than its baseline, occur at moderate pruning levels. For example, DINOv2 with 8 removed MLP sublayers runs $1.14\times$ faster and SwinV2 with 8 removed blocks $1.33\times$ faster, both with $\gamma < 0$. Joint Layer Drop keeps $\gamma$ negative over the widest range, up to 40 removals on DINOv2 and 16 on SwinV2, whereas Block Drop turns positive from 8 removals on DINOv2 and at 16 on SwinV2. At extreme drop counts, $\gamma$ ceases to be informative. DINOv2 with 40 removed blocks reaches only 27.7\% average accuracy, yet its $74\times$ speedup keeps $\gamma$ close to zero, and ViT with 12 removed blocks behaves the same way. The ratio is therefore meaningful only at moderate pruning, where accuracy remains close to the baseline.

During the throughput measurements we also sample the board power of the GPU through NVML every 50~ms and report the energy per image as the average power divided by the throughput. The A40 draws close to its 300~W power limit at moderate drop counts, so the energy per image falls in proportion to the throughput gain. Removing 8 blocks from DINOv2 lowers it from 2.20 to 1.77~J per image, a 19.5\% saving, while the configurations of DINOv2 and SwinV2 that remain equivalent to the baseline under Joint Layer Drop save between 3.2\% and 16.7\%.

\subsection{Automated Model Selection}
\label{subsec:model_selection}
The experiments presented in the previous sections demonstrate that pruning effectiveness
depends on the interplay between architecture, pruning strategy, number of removed layers
and target dataset. No single configuration dominates universally, as the optimal choice
varies with dataset complexity, class structure and domain characteristics. A practitioner seeking to deploy a pruned model on a new dataset therefore faces a large,
complex search space, and the best choice differs from one deployment scenario to the next.
Ideally, one wants the most accurate model with the fastest inference and lowest computational
cost. In our setting, 4 architectures, 4 pruning strategies, and up to 40 drop counts
yield hundreds of candidate configurations. Evaluating each one manually is prohibitively
expensive, yet a suboptimal choice either leaves performance on the table or sacrifices more
speed than necessary. 

To address this, we propose an automated
model selection pipeline that, given a target dataset, systematically explores the
configuration space and returns a single deployment-ready model, namely the pruned
configuration with the highest validation accuracy within a predefined tolerance of the
unpruned baseline.
\begin{algorithm}[t]
\small
\caption{Automated Model Selection}
\small
\label{alg:model_selection}
\begin{algorithmic}[1]
\REQUIRE Dataset $\mathcal{D}$, architectures $\mathcal{A}$, methods $\mathcal{M}$, step $\Delta$, threshold $\tau$
\ENSURE $(a^*, m^*, n^*)$, test metrics
\STATE \COMMENT{Phase 1: Baseline}
\FOR{$(a, m) \in \mathcal{A} \times \mathcal{M}$}
    \STATE $\text{model} \leftarrow \textsc{LoadPruned}(a, m, 0)$
    \STATE $\text{model} \leftarrow \textsc{FinetuneHead}(\text{model}, \mathcal{D}_{\text{train}}, \text{epochs}=5)$
    \STATE $\text{Acc}_{\text{base}}(a, m) \leftarrow \textsc{Evaluate}(\text{model}, \mathcal{D}_{\text{val}})$
\ENDFOR
\STATE \COMMENT{Phase 2: Search}
\FOR{$(a, m) \in \mathcal{A} \times \mathcal{M}$}
    \FOR{$n = \Delta, 2\Delta, \ldots, n_{\max}(a)$}
        \STATE $\text{model} \leftarrow \textsc{LoadPruned}(a, m, n)$
        \STATE $\text{model} \leftarrow \textsc{FinetuneHead}(\text{model}, \mathcal{D}_{\text{train}}, \text{epochs}=5)$
        \STATE $\text{Acc}_n \leftarrow \textsc{Evaluate}(\text{model}, \mathcal{D}_{\text{val}})$
        \IF{$\text{Acc}_{\text{base}}(a, m) - \text{Acc}_n > \tau$}
            \STATE \textbf{break}
        \ENDIF
    \ENDFOR
\ENDFOR
\STATE \COMMENT{Phase 3: Selection and Final Head Training}
\STATE $(a^*, m^*, n^*) \leftarrow \arg\max_{(a,m,n)} \text{Acc}_n$ \COMMENT{early-stopped configurations excluded; ties: prefer larger $n$}
\STATE $\text{model} \leftarrow \textsc{LoadPruned}(a^*, m^*, n^*)$
\STATE $\text{model} \leftarrow \textsc{FinetuneHead}(\text{model}, \mathcal{D}_{\text{train}} \cup \mathcal{D}_{\text{val}}, \text{epochs}=20)$
\STATE $\text{metrics} \leftarrow \textsc{Evaluate}(\text{model}, \mathcal{D}_{\text{test}})$
\RETURN $(a^*, m^*, n^*)$, metrics
\end{algorithmic}
\end{algorithm}
The pipeline operates in 3 sequential phases, summarized in Algorithm~\ref{alg:model_selection}. Throughout, only the classification head is trained while the backbone remains frozen, so that the evaluation reflects the quality of the pruned representations rather than the capacity to recover through extensive retraining. All head training uses AdamW (learning rate $0.001$, weight decay $0.03$) with a cosine annealing schedule and a batch size of 32. In the first phase, the unpruned model is trained for 5 epochs on the training split and evaluated on the validation split, giving the reference accuracy $\text{Acc}_{\text{base}}$. In the second phase, for each pruning method the pipeline iterates over increasing drop counts in steps of $\Delta = 4$, training each pruned model for 5 epochs and evaluating it on the validation split. If the accuracy $\text{Acc}_{\text{drop}}$ falls more than $\tau$ below $\text{Acc}_{\text{base}}$, with accuracies as fractions in $[0,1]$ and $\tau = 0.05$, the search along that method stops and moves to the next one. This avoids evaluating deeply pruned configurations whose speedup is not worth the accuracy lost. Both $\tau$ and $\Delta$ are configurable; reducing $\Delta$ increases search granularity at the cost of more evaluations, while lowering $\tau$ tightens the quality requirement. In the third phase, the configuration with the highest validation accuracy is selected, with ties broken in favor of the higher drop count since it provides greater compression, and is trained for 20 epochs on the combined training and validation splits with linear warmup, then evaluated once on the held-out test split to produce the final deployment-ready model.

\begin{table}[t]
\centering
\caption{Model selection results on the 12 downstream datasets. For each architecture, Unpruned is the test accuracy of the unpruned model and Selected the test accuracy of the configuration selected on the validation split, both trained with the third-phase protocol, followed by the method, drop count and FLOPs reduction of the selected configuration. Baseline marks datasets on which no pruned configuration exceeded the unpruned model on the validation split, and $\dagger$ marks the model returned by the full pipeline.}
\label{tab:best_results_per_arch}
\resizebox{\textwidth}{!}{%
\begin{tabular}{l|ccccc|ccccc|ccccc|ccccc}
\toprule
\multirow{2}{*}{\textbf{Dataset}} & \multicolumn{5}{c|}{\textbf{DINOv2}} & \multicolumn{5}{c|}{\textbf{DINOv3}} & \multicolumn{5}{c|}{\textbf{SwinV2}} & \multicolumn{5}{c}{\textbf{ViT}} \\
\cmidrule(lr){2-6} \cmidrule(lr){7-11} \cmidrule(lr){12-16} \cmidrule(lr){17-21}
 & \textbf{Unpruned (\%)} & \textbf{Selected (\%)} & \textbf{Method} & \textbf{Drop} & \textbf{FLOPs$\downarrow$ (\%)} & \textbf{Unpruned (\%)} & \textbf{Selected (\%)} & \textbf{Method} & \textbf{Drop} & \textbf{FLOPs$\downarrow$ (\%)} & \textbf{Unpruned (\%)} & \textbf{Selected (\%)} & \textbf{Method} & \textbf{Drop} & \textbf{FLOPs$\downarrow$ (\%)} & \textbf{Unpruned (\%)} & \textbf{Selected (\%)} & \textbf{Method} & \textbf{Drop} & \textbf{FLOPs$\downarrow$ (\%)} \\
\midrule
CIFAR-10     & 99.32 & 99.30 & Block$^\dagger$ & 4 & 9.98 & 99.14 & 99.14 & baseline & 0 & 0.00 & 97.38 & 97.38 & baseline & 0 & 0.00 & 93.75 & 93.75 & baseline & 0 & 0.00 \\
Zoolake      & 95.77 & 96.06 & MLP$^\dagger$ & 8 & 12.94 & 95.62 & 95.62 & baseline & 0 & 0.00 & 93.57 & 94.17 & MLP & 4 & 10.31 & 93.31 & 93.68 & MLP & 4 & 20.99 \\
Laryngeal    & 98.48 & 99.24 & Joint$^\dagger$ & 24 & 28.45 & 97.35 & 97.35 & baseline & 0 & 0.00 & 95.83 & 98.48 & MLP & 8 & 20.62 & 91.29 & 95.45 & Block & 4 & 32.83 \\
Pest         & 98.90 & 97.24 & Joint$^\dagger$ & 16 & 16.25 & 97.51 & 97.51 & baseline & 0 & 0.00 & 87.85 & 93.37 & MLP & 8 & 20.62 & 95.86 & 95.86 & baseline & 0 & 0.00 \\
InfLarynge   & 91.67 & 96.25 & Joint$^\dagger$ & 40 & 50.64 & 93.33 & 95.00 & Joint & 20 & 42.78 & 95.83 & 96.25 & Joint & 24 & 46.07 & 95.42 & 95.42 & MLP & 4 & 20.99 \\
WHOI22       & 94.26 & 94.62 & MLP$^\dagger$ & 8 & 12.94 & 94.65 & 94.65 & baseline & 0 & 0.00 & 93.96 & 93.96 & baseline & 0 & 0.00 & 92.22 & 91.95 & Block & 4 & 32.83 \\
Kaggle38     & 92.39 & 91.49 & Attn & 8 & 7.01 & 92.29 & 92.29 & baseline & 0 & 0.00 & 93.17 & 92.22 & Attn$^\dagger$ & 4 & 4.83 & 89.10 & 89.58 & MLP & 4 & 20.99 \\
ZooScan20    & 82.76 & 84.06 & MLP$^\dagger$ & 8 & 12.94 & 81.09 & 81.09 & baseline & 0 & 0.00 & 84.43 & 84.43 & baseline & 0 & 0.00 & 76.04 & 78.54 & MLP & 4 & 20.99 \\
Bark         & 90.69 & 89.75 & Joint$^\dagger$ & 40 & 50.64 & 88.64 & 88.64 & baseline & 0 & 0.00 & 86.54 & 87.07 & MLP & 12 & 30.93 & 80.81 & 83.85 & Block & 4 & 32.83 \\
CrossD       & 54.32 & 60.39 & MLP$^\dagger$ & 20 & 32.36 & 68.56 & 60.93 & Joint & 8 & 17.84 & 66.82 & 54.15 & MLP & 8 & 20.62 & 32.66 & 27.79 & MLP & 4 & 20.99 \\
LCZ42        & 63.81 & 61.22 & Joint$^\dagger$ & 28 & 33.44 & 63.28 & 65.27 & Joint & 20 & 42.78 & 62.05 & 61.72 & MLP & 12 & 30.93 & 48.15 & 41.27 & Attn & 8 & 23.68 \\
ColorBG      & 91.67 & 96.67 & Joint & 40 & 50.64 & 90.00 & 90.00 & Block & 12 & 49.87 & 95.00 & 98.33 & Block & 4 & 16.75 & 95.00 & 95.00 & MLP$^\dagger$ & 4 & 20.99 \\
\midrule
\textbf{Mean}& \textbf{87.84} & \textbf{88.86} & -- & \textbf{20.33} & \textbf{26.52} & \textbf{88.45} & \textbf{88.12} & -- & \textbf{5.00} & \textbf{12.77} & \textbf{87.70} & \textbf{87.63} & -- & \textbf{7.00} & \textbf{16.81} & \textbf{81.97} & \textbf{81.84} & -- & \textbf{3.67} & \textbf{20.68} \\
\bottomrule
\end{tabular}%
}
\end{table}

Table~\ref{tab:best_results_per_arch} presents our findings. For each dataset, it reports the best configuration of each architecture, obtained by applying the selection rule to that architecture alone, and marks with $\dagger$ the single model returned by the full pipeline. In this per-architecture view the unpruned model is also a candidate, and it is marked as baseline when no pruned variant exceeds it on the validation split. Every configuration is trained for 20 epochs on the combined training and validation splits and evaluated once on the test split, following the third phase of the pipeline. The unpruned column reports the unpruned model trained and evaluated with the same protocol, so that the 2 accuracy columns differ only in the pruning configuration. Because of this protocol, the accuracies here are higher than those of Section~\ref{subsec:finetuning}, which used 5 epochs on the training split alone. The table should be read per architecture rather than as a cross-architecture ranking.

The selected drop count ranges from 0 to 40 across datasets, which confirms that the best operating point is dataset dependent. DINOv3 keeps the unpruned model on 8 of the 12 datasets, consistent with its fragility under depth removal. Since the selection is made on validation accuracy after 5 epochs while the table reports test accuracy after 20 epochs, a selected configuration can score below its baseline on test. On the in-domain datasets these losses are small, below 2 points, as on Pest, Kaggle38 and Bark for DINOv2 or Kaggle38 for SwinV2. The larger losses concentrate on the 2 cross-domain datasets, CrossD and LCZ42, where the validation and test splits come from different instruments or regions, so a configuration selected on validation accuracy has no guarantee of transferring to the test distribution. Averaged over the 12 datasets, test accuracy changes by $+1.0$, $-0.3$, $-0.1$ and $-0.1$ points for DINOv2, DINOv3, SwinV2 and ViT, while FLOPs decrease by 27\%, 13\%, 17\% and 21\%.  Pruning therefore delivers the computational saving anticipated by the previous sections at an average accuracy cost close to zero, with the largest individual losses appearing under distribution shift, where validation accuracy is a weaker guide to test accuracy.

\subsection{Stacking Depth Pruning with Weight Quantization}
\label{subsec:quantization}
The efficiency analysis of Section~\ref{subsec:efficiency} shows that depth pruning reduces computation and increases throughput, but it reduces memory only in proportion to the removed depth, whereas weight quantization compresses the stored parameters without touching the depth of the network. The 2 techniques are therefore orthogonal in principle, and in this section we test whether they compose in practice, that is, whether a model that has already been depth-pruned tolerates quantization as well as the unpruned one does.

We select 11 configurations from Table~\ref{tab:best_results_per_arch}, covering all 4 architectures and all 4 pruning strategies. ColorBG is excluded, since each prediction on its 60-image test split shifts accuracy by 1.7 points. For each configuration, and for the corresponding unpruned model, we train the linear classification head once in FP32 with the 5-epoch protocol of Section~\ref{subsec:finetuning}. We then attach this same head to 4 copies of the backbone, one kept in FP32, one cast to BF16, and 2 quantized post-training to 8-bit and 4-bit weights using the LLM.int8 and NF4 schemes of bitsandbytes \citep{dettmers2022llmint8, dettmers2023qlora}. Quantization is weight-only and requires no calibration data or gradient step, so no training is involved beyond the linear head. Since the head is shared across the 4 copies, any difference in test accuracy or in the macro-averaged metrics is attributable solely to the quantization of the backbone. Table~\ref{tab:quant_stack} reports one representative configuration per architecture, with the weight footprint and the throughput measured at batch size 1, the regime of interest for edge deployment where weights dominate memory. The complete results for every configuration are reported in Appendix~\ref{app:quantization}. Because the heads are trained with the 5-epoch protocol, absolute accuracies differ from the 20-epoch values of Table~\ref{tab:best_results_per_arch}. 
\begin{center}
\captionof{table}{Stacking depth pruning with weight 
quantization, one representative configuration per architecture. Each block
reports the unpruned FP32 model followed by the pruned model at 4
precisions, all sharing the same classification head. Reduction is the weight
footprint and Speedup the throughput, both relative to the unpruned FP32 model
and measured at batch size 1 on a single NVIDIA A40.}
\label{tab:quant_stack}
{\small
\begin{tabular}{lllrrrrr}
\toprule
\textbf{Dataset / Model} & \textbf{Pruning} & \textbf{Precision} & \textbf{Acc (\%)} & \textbf{F1} & \textbf{Weights (MB)} & \textbf{Reduction} & \textbf{Speedup} \\
\midrule
\multirow{5}{*}{Laryngeal / DINOv2} & baseline & fp32 & 98.48 & 0.985 & 4,335 & 1.0$\times$ & 1.00$\times$ \\
\cmidrule(lr){2-8}
 & \multirow{4}{*}{Joint-24} & fp32 & 99.24 & 0.992 & 3,110 & 1.4$\times$ & 1.44$\times$ \\
 &  & bf16 & 99.24 & 0.992 & 1,555 & 2.8$\times$ & 4.85$\times$ \\
 &  & int8 & 98.86 & 0.989 & 781 & 5.6$\times$ & 2.21$\times$ \\
 &  & int4 & 99.24 & 0.992 & 394 & 11.0$\times$ & 4.47$\times$ \\
\midrule
\multirow{5}{*}{Bark / SwinV2} & baseline & fp32 & 82.27 & 0.747 & 343 & 1.0$\times$ & 1.00$\times$ \\
\cmidrule(lr){2-8}
 & \multirow{4}{*}{MLP-12} & fp32 & 86.88 & 0.819 & 223 & 1.5$\times$ & 1.05$\times$ \\
 &  & bf16 & 86.80 & 0.817 & 117 & 2.9$\times$ & 1.04$\times$ \\
 &  & int8 & 86.96 & 0.820 & 64 & 5.3$\times$ & 0.28$\times$ \\
 &  & int4 & 86.19 & 0.813 & 38 & 9.1$\times$ & 0.48$\times$ \\
\midrule
\multirow{5}{*}{Zoolake / ViT} & baseline & fp32 & 91.53 & 0.752 & 327 & 1.0$\times$ & 1.00$\times$ \\
\cmidrule(lr){2-8}
 & \multirow{4}{*}{MLP-4} & fp32 & 92.16 & 0.735 & 255 & 1.3$\times$ & 1.08$\times$ \\
 &  & bf16 & 92.12 & 0.736 & 128 & 2.6$\times$ & 1.08$\times$ \\
 &  & int8 & 92.20 & 0.734 & 65 & 5.1$\times$ & 0.18$\times$ \\
 &  & int4 & 90.94 & 0.735 & 33 & 9.9$\times$ & 0.32$\times$ \\
\midrule
\multirow{5}{*}{LCZ42 / DINOv3} & baseline & fp32 & 63.65 & 0.505 & 1,156 & 1.0$\times$ & 1.00$\times$ \\
\cmidrule(lr){2-8}
 & \multirow{4}{*}{Joint-20} & fp32 & 65.00 & 0.513 & 660 & 1.8$\times$ & 1.67$\times$ \\
 &  & bf16 & 64.10 & 0.508 & 330 & 3.5$\times$ & 1.76$\times$ \\
 &  & int8 & 3.75 & 0.004 & 166 & 7.0$\times$ & 0.31$\times$ \\
 &  & int4 & 1.10 & 0.001 & 84 & 13.8$\times$ & 0.67$\times$ \\
\bottomrule
\end{tabular}}
\end{center}

Regarding memory, the 2 reductions compound multiplicatively. On Laryngeal, Joint Layer Drop with 24 removals followed by 4-bit quantization shrinks the DINOv2 weights from 4,335~MB to 394~MB, an 11$\times$ reduction, of which depth pruning contributes a factor of 1.4 and quantization the remaining 7.9. With 40 removals, as selected on Bark and InfLarynge, the combined reduction reaches 15.9$\times$ (Appendix~\ref{app:quantization}). Regarding accuracy, DINOv2 is insensitive to quantization whether pruned or not, with 4-bit accuracy and macro F1-score within 0.6 points of FP32 on all 6 configurations. On SwinV2 and ViT, 8-bit weights stay within 0.3 points of FP32 except on LCZ42, where they cost 1.6 to 1.7 points and BF16 alone already costs 1.1 to 1.2. With 4-bit weights the loss reaches 1.2 points on Bark, Kaggle38 and Zoolake and 5.3 points on LCZ42, where the unpruned SwinV2 loses 10.6 points under the same quantization. DINOv3 stands apart once more. Both quantized variants collapse to predicting a single class, with a macro recall of exactly $1/17$ on the 17 LCZ42 classes, while the BF16 variant is unaffected. The collapse occurs identically on the pruned and unpruned models, so it is a property of the backbone rather than of pruning, and we conjecture that the activation statistics of DINOv3 are not handled by the outlier treatment of the bitsandbytes schemes. Regarding throughput, quantization saves memory but not time. With the bitsandbytes kernels used here, 8-bit and 4-bit inference are slower than BF16 on every backbone, so BF16 remains the fastest precision and lower bit widths trade speed for memory. Taken together, the pruned configurations of Table~\ref{tab:best_results_per_arch} can be quantized to 4 bits on DINOv2 and to 8 bits on SwinV2 and ViT with no meaningful accuracy cost outside LCZ42, for a further memory reduction of 4$\times$ to 8$\times$, while DINOv3 requires a quantization scheme other than the ones tested here.

\subsection{Comparison with Related Structural Compression Approaches}
\label{subsec:comparison}
We close the experimental analysis by comparing our results with the structural compression approaches of Section~\ref{sec:related_work}. Most of them cannot be compared with ours, since they either recover accuracy through training or publish no operating point commensurable with ours, which leaves CORP and TOAST. CORP (\cite{zhang2026corpclosedformoneshotrepresentationpreserving}) differs from our method in 4 respects. It narrows widths instead of removing depth, compensates for the removed structures with a closed-form correction of the remaining weights and is evaluated on ImageNet-1K alone. It also refits this correction at every sparsity level on 4,000 calibration images, whereas our ranking is computed once on 512 images and reused for every drop count. Since its results come from a different protocol, we use the values published for the shared DeiT-Base checkpoint to contextualize the achievable computational reduction rather than to rank the methods. TOAST (\cite{cannistraci2026toast}) instead reduces depth like our Block Drop, replacing contiguous runs of whole blocks with a fitted linear map, which allows us to rerun it under our protocol on ViT-Base, DeiT-Base and DINOv2-Giant. We use the authors' code, which with their preprocessing reproduces their published single-block results on ViT-Base and DeiT-Base, matching the unpruned accuracy and the cost of the most replaceable block within 0.01 points. The linear maps are fitted on 500 ImageNet-1K training images, and the contiguous run of blocks to replace is selected either by the lowest linear approximation error on these images or by the highest accuracy on 2,000 further training images. We compute this error over all tokens, because the original error on the CLS tokens of 50 images is zero for every run at these hidden sizes.

\begin{center}
\centering
\small
\captionof{table}{Comparison with CORP on DeiT-Base, evaluated on the full ImageNet-1K validation set with the original classifier. FLOPs reductions are relative to each work's own dense count.}
\label{tab:sota_comparison}
\begin{tabular}{c ccc ccc}
\toprule
 & \multicolumn{3}{c}{\textbf{CORP} (closed-form compensation)} & \multicolumn{3}{c}{\textbf{Block Drop} (no compensation)} \\
\cmidrule(lr){2-4} \cmidrule(lr){5-7}
 & FLOPs$\downarrow$ (\%) & Speedup & Top-1 (\%) & FLOPs$\downarrow$ (\%) & Speedup & Top-1 (\%) \\
\midrule
Baseline & 0.0  & $1.00\times$ & 81.74 & 0.0  & $1.00\times$ & 81.81 \\
$n = 4$  & 35.0 & $1.31\times$ & 75.03 & 32.8 & $1.44\times$ & 72.16 \\
$n = 5$  & 43.0 & $1.51\times$ & 72.00 & 41.0 & $1.62\times$ & 67.77 \\
\bottomrule
\end{tabular}
\end{center}

On DeiT-Base, the compact backbone that our analysis predicts to be the least favorable case, Block Drop stays within 3 points of CORP at a comparable budget and 4.2 points behind it at about 42\%, a gap that measures the value of CORP's compensation. Since our method is orthogonal to compensation, quantization and lightweight tuning, all of them can be applied on top of the model it outputs, as Section~\ref{subsec:quantization} showed for quantization.

All differences between Block Drop and TOAST in Table~\ref{tab:toast_comparison} are significant under a paired McNemar test (\cite{mcnemar1947note}) with Holm correction ($p < 0.001$), except on DINOv2-Giant, where Block Drop does not differ from the validation-selected variant up to 8 removed blocks, nor from either variant at 2.

The outcome depends on the backbone. On ViT-Base, the better TOAST variant is more accurate than Block Drop at every budget, and computing our ranking on the TOAST images raises Block Drop by up to 7.0 points. Removing MLP sublayers instead of whole blocks closes the gap at the smallest budget, since Joint Layer Drop reaches 77.98\% at a 10.5\% FLOPs reduction, within 0.1 points of the best TOAST variant at 7.6\%. On DeiT-Base, Block Drop is more accurate at every budget, by 13.4 points over the best TOAST variant at 4 removed blocks. From 3 removed blocks, the blocks selected by our ranking are not contiguous, namely 2, 3 and 11, then 2, 3, 9 and 11, whereas TOAST can only replace contiguous runs. On DINOv2-Giant, the validation-selected TOAST retains 3.0 and 9.3 more points than Block Drop at 12 and 16 removed blocks, although our ranking computed on the TOAST images exceeds it at 12. Selecting the run by approximation error instead of accuracy degrades TOAST at larger budgets, down to 13.56\% on DINOv2-Giant at 16 removed blocks. Since the linear map costs a single matrix product, the measured speedups of the 2 methods differ by at most 1\% at equal $n$.

In summary, our method reaches FLOPs reductions comparable to CORP with larger measured speedups and a far cheaper selection procedure, at a larger accuracy cost on DeiT-Base. Against TOAST at an equal number of removed blocks, it is more accurate on DeiT-Base, less accurate on ViT-Base and not significantly different on DINOv2-Giant up to a fifth of its depth.

\begin{center}
\centering
\small
\captionof{table}{Comparison with TOAST on the full ImageNet-1K validation set with the original classifier, at an equal number $n$ of removed blocks. Block Drop ranks the blocks on $\mathcal{D}_{\text{calib}}$, and Block Drop$^{\dagger}$ on the 500 training images on which TOAST fits its linear maps. TOAST (error) replaces the contiguous run with the lowest linear approximation error, and TOAST (val.) the run with the highest accuracy on 2,000 further training images. FLOPs reductions are those of Block Drop, while TOAST removes 0.7 percentage points less on ViT-Base and DeiT-Base and 0.2 less on DINOv2-Giant because of its linear map. Bold marks the most accurate method at each $n$.}
\label{tab:toast_comparison}
\begin{tabular}{l c c c c c c}
\toprule
\textbf{Model} & $n$ & \textbf{FLOPs$\downarrow$ (\%)} & \textbf{Block Drop} & \textbf{Block Drop$^{\dagger}$} & \textbf{TOAST (error)} & \textbf{TOAST (val.)} \\
\midrule
\multirow{5}{*}{ViT-Base}
             & 0  & 0.0  & \multicolumn{4}{c}{80.32} \\
\cmidrule(lr){4-7}
             & 1  & 8.2  & 73.22 & 76.05 & 77.09 & \textbf{78.07} \\
             & 2  & 16.4 & 59.61 & 66.58 & 65.25 & \textbf{73.05} \\
             & 3  & 24.6 & 36.94 & 36.94 & 48.99 & \textbf{53.32} \\
             & 4  & 32.8 & 3.25  & 9.66  & \textbf{21.81} & 21.80 \\
\midrule
\multirow{7}{*}{DeiT-Base}
             & 0  & 0.0  & \multicolumn{4}{c}{81.81} \\
\cmidrule(lr){4-7}
             & 1  & 8.2  & \textbf{81.37} & \textbf{81.37} & 79.80 & 80.94 \\
             & 2  & 16.4 & \textbf{79.19} & \textbf{79.19} & 77.97 & 78.48 \\
             & 3  & 24.6 & \textbf{77.45} & \textbf{77.45} & 71.08 & 71.59 \\
             & 4  & 32.8 & \textbf{72.16} & 70.09 & 14.11 & 58.79 \\
             & 5  & 41.0 & \textbf{67.77} & 54.37 & 2.89  & 36.58 \\
             & 6  & 49.2 & \textbf{38.52} & 36.95 & 1.67  & 14.57 \\
\midrule
\multirow{6}{*}{DINOv2-Giant}
             & 0  & 0.0  & \multicolumn{4}{c}{86.53} \\
\cmidrule(lr){4-7}
             & 2  & 5.0  & 86.52 & 86.52 & 86.47 & \textbf{86.53} \\
             & 4  & 10.0 & \textbf{86.46} & 86.45 & 86.19 & 86.36 \\
             & 8  & 20.0 & 85.67 & \textbf{85.81} & 82.94 & 85.49 \\
             & 12 & 29.9 & 81.08 & \textbf{84.65} & 62.66 & 84.03 \\
             & 16 & 39.9 & 65.07 & 65.07 & 13.56 & \textbf{74.38} \\
\bottomrule
\end{tabular}
\end{center}

\section{Conclusion}
\label{sec:conclusion}
In this paper, we presented a systematic study of layer-level structural redundancy in vision transformers, extending similarity-based pruning, originally developed for LLMs, to diverse computer vision architectures without retraining their backbones. By evaluating cosine similarity between sublayer inputs and outputs across flat and hierarchical backbones, we demonstrated that depth redundancy is a prevalent property of several deep vision transformers.

Our empirical evaluation across 13 diverse benchmarks spanning natural images, remote sensing, medical imaging, microscopy, and agriculture yields several key insights:
\begin{enumerate}
    \item {Intrinsic Layer Redundancy:} Deep self-supervised models such as DINOv2 exhibit substantial architectural redundancy, allowing 20\% of transformer blocks to be removed without any retraining and with minimal performance impact (e.g., achieving a 24\% throughput increase and a 20\% lower energy per image with less than a 1\% relative accuracy reduction). 
    \item {Stability of Joint Sublayer Pruning:} With a fine-tuned head, Joint Layer Drop produces the most gradual median degradation on every architecture, because ranking attention and MLP sublayers together lets it switch to the other sublayer type before the collapse that single-type removal eventually reaches.
    \item {Lightweight Adaptation:} Combining structural pruning with lightweight classification head fine-tuning allows pruned variants to maintain, and in several cross-domain and specialized tasks even surpass, the baseline performance of full-depth architectures, all while significantly reducing FLOPs and increasing inference throughput.
    \item {Automated Deployment Pipeline:} To eliminate the combinatorial burden of manual model selection, we introduced an automated three-phase search framework with early stopping. This pipeline identifies, for a given downstream dataset, the pruned configuration with the highest validation accuracy within a predefined tolerance.
    \item {Composability with Quantization:} Depth pruning is orthogonal to weight quantization, and the two compound multiplicatively. Pruned DINOv2 models tolerate 4-bit weights and pruned SwinV2 and ViT models 8-bit weights without meaningful accuracy loss, for a combined weight reduction of up to 16$\times$.
    \item {Divergence from LLM Pruning Behavior:} In the evaluation without
    retraining, we observe redundancy dynamics that differ from those
    reported for LLMs, with block removal less harmful than sublayer removal on DINOv2, a more redundant sublayer type that changes from one backbone to another, and Joint Layer Drop benefiting mainly the hierarchical architecture, suggesting that pruning guidelines derived from LLMs should
    not be assumed to transfer to vision backbones.
\end{enumerate}

These findings are relevant to expert systems that must run pretrained vision backbones under a fixed computational budget, such as the endoscopic, histopathological and plankton monitoring applications represented in our benchmarks, where retraining a backbone is often not an option.

For practitioners, our results translate into a set of simple operating guidelines. Deep backbones such as DINOv2 tolerate depth pruning best. On DINOv2, Block Drop up to 8 removals and Joint Layer Drop up to 12 removals keep accuracy equivalent to the baseline within 2 points, with speedups of up to $1.24\times$ and $1.12\times$ respectively. On the hierarchical SwinV2, Attention and Joint Layer Drop remain equivalent up to 8 removals, with a $1.21\times$ speedup, and the evidence becomes inconclusive beyond that. Compact backbones such as ViT-Base should be pruned only lightly. At 4 removals no strategy differs significantly from the baseline, but none reaches equivalence either, and from 8 removals Attention Drop already degrades accuracy. DINOv3 should not be depth-pruned without a recovery mechanism, as every configuration we tested degrades significantly. When the best configuration is not obvious a priori, which our results show is the common case, the automated selection pipeline resolves the choice at the cost of a single search run, with the caveat that its validation-based selection is a weaker guide under distribution shift, where the largest test losses occurred. Once a pruned model is selected, its weights can be quantized post-training to 4 bits on DINOv2 and to 8 bits on SwinV2 and ViT.

Several promising avenues remain open for future investigation:
\begin{itemize}
\item {Hybrid and Complementary Compression:} Combining layer-level depth
    reduction with spatial token-level sparsification (e.g., token merging or
    dynamic pruning) to achieve combined speedups across both network depth
    and sequence length. Beyond the weight quantization studied here, structural layer pruning could also be combined with low-rank factorization and with quantization schemes that keep the throughput of BF16, further reducing memory footprint, latency and energy consumption.
    \item {Advanced Fine-Tuning Protocols:} Exploring parameter-efficient fine-tuning strategies, such as LoRA or adapter modules, beyond frozen-backbone linear head adaptation to investigate whether light backbone updates can recover accuracy under extreme pruning ratios ($>50\%$). Moreover, we plan to investigate further fine-tuning of layers targeted for pruning, incorporating a loss function that explicitly accounts for the cosine distance (or other similarity metrics) between layers.
    \item {Extension to Dense Prediction and Multimodal Architectures:} Extending the similarity-based layer dropping methodology to object detection, semantic segmentation, and multimodal vision-language backbones.
\end{itemize}

\bibliographystyle{plainnat}
\bibliography{references}

\clearpage
\appendix
\addtocontents{toc}{\protect\setcounter{tocdepth}{1}}
\section*{\LARGE Appendix}
\makeatletter
\@starttoc{toc}
\makeatother

\newpage
\section{Per-Layer Calibration Similarities}
\label{app:similarities}
\begin{center}
\captionof{table}{Exact per-layer calibration cosine similarities for the 5
evaluated backbones summarized in Figure~\ref{fig:vision_similarities}. Block denotes the
similarity between the full block input and output, while Attn and MLP denote
the sublayer-level similarities. Bold values indicate the highest similarity
within each column, and dashes mark layers beyond an architecture's depth.}
\label{tab:sim_all}
{\scriptsize
\setlength{\tabcolsep}{3pt}
\renewcommand{\arraystretch}{1}
\begin{tabular}{c|ccc|ccc|ccc|ccc|ccc}
\toprule
\multirow{2}{*}{\textbf{Layer}} & \multicolumn{3}{c|}{\textbf{ViT}} & \multicolumn{3}{c|}{\textbf{DeiT}} & \multicolumn{3}{c|}{\textbf{DINOv2}} & \multicolumn{3}{c|}{\textbf{DINOv3}} & \multicolumn{3}{c}{\textbf{SwinV2}} \\
\cmidrule(lr){2-4}\cmidrule(lr){5-7}\cmidrule(lr){8-10}\cmidrule(lr){11-13}\cmidrule(lr){14-16}
 & Block & Attn & MLP & Block & Attn & MLP & Block & Attn & MLP & Block & Attn & MLP & Block & Attn & MLP \\
\midrule
0 & 0.625 & 0.871 & 0.921 & 0.593 & 0.825 & 0.895 & 0.939 & 0.521 & 0.120 & 0.705 & 0.774 & 0.655 & 0.235 & -0.110 & 0.433 \\
1 & 0.777 & 0.890 & 0.963 & 0.848 & 0.869 & 0.970 & 0.977 & 0.593 & 0.231 & 0.943 & 0.795 & 0.708 & 0.711 & 0.744 & 0.809 \\
2 & 0.781 & 0.876 & 0.962 & \textbf{0.897} & 0.907 & 0.976 & 0.982 & 0.571 & 0.258 & 0.932 & 0.793 & 0.738 & 0.637 & 0.794 & 0.746 \\
3 & 0.822 & 0.903 & 0.968 & 0.885 & 0.924 & 0.975 & 0.981 & 0.500 & 0.334 & \textbf{0.970} & 0.782 & 0.750 & 0.810 & 0.898 & 0.878 \\
4 & 0.841 & 0.919 & 0.965 & 0.849 & 0.899 & 0.969 & 0.981 & 0.369 & 0.382 & 0.924 & 0.653 & 0.605 & 0.811 & 0.848 & 0.903 \\
5 & 0.866 & 0.933 & 0.974 & 0.853 & 0.904 & 0.969 & 0.981 & 0.369 & 0.375 & 0.818 & 0.501 & 0.522 & 0.872 & 0.902 & 0.922 \\
6 & 0.877 & 0.939 & 0.979 & 0.852 & 0.912 & 0.970 & \textbf{0.984} & 0.478 & 0.448 & 0.889 & 0.720 & 0.658 & 0.890 & 0.932 & 0.929 \\
7 & 0.883 & 0.946 & 0.981 & 0.863 & 0.930 & 0.973 & 0.981 & 0.352 & 0.534 & 0.929 & 0.794 & 0.759 & 0.908 & 0.948 & 0.932 \\
8 & \textbf{0.884} & 0.950 & 0.980 & 0.860 & 0.944 & 0.972 & 0.982 & 0.425 & 0.594 & 0.934 & 0.808 & 0.828 & 0.920 & 0.952 & 0.946 \\
9 & 0.870 & \textbf{0.953} & 0.974 & 0.867 & \textbf{0.957} & 0.971 & 0.980 & 0.416 & 0.619 & 0.951 & 0.865 & 0.879 & 0.934 & 0.956 & 0.957 \\
10 & 0.766 & 0.947 & 0.928 & 0.829 & 0.948 & 0.958 & 0.978 & 0.386 & 0.646 & 0.953 & 0.836 & 0.907 & 0.945 & 0.969 & 0.962 \\
11 & 0.806 & 0.865 & \textbf{0.984} & 0.874 & 0.867 & \textbf{0.994} & 0.977 & 0.410 & 0.663 & 0.952 & 0.950 & 0.939 & 0.953 & 0.974 & 0.971 \\
12 & -- & -- & -- & -- & -- & -- & 0.975 & 0.402 & 0.674 & 0.928 & 0.924 & 0.950 & 0.960 & 0.980 & 0.974 \\
13 & -- & -- & -- & -- & -- & -- & 0.974 & 0.447 & 0.702 & 0.920 & 0.941 & 0.962 & 0.972 & 0.989 & 0.980 \\
14 & -- & -- & -- & -- & -- & -- & 0.972 & 0.490 & 0.711 & 0.900 & 0.961 & 0.976 & 0.974 & 0.989 & 0.983 \\
15 & -- & -- & -- & -- & -- & -- & 0.971 & 0.506 & 0.738 & 0.904 & 0.971 & 0.981 & 0.982 & 0.994 & 0.987 \\
16 & -- & -- & -- & -- & -- & -- & 0.970 & 0.576 & 0.760 & 0.892 & 0.981 & 0.984 & \textbf{0.986} & 0.995 & 0.990 \\
17 & -- & -- & -- & -- & -- & -- & 0.969 & 0.657 & 0.771 & 0.872 & 0.985 & 0.991 & \textbf{0.986} & 0.994 & \textbf{0.991} \\
18 & -- & -- & -- & -- & -- & -- & 0.968 & 0.665 & 0.794 & 0.856 & 0.988 & 0.995 & 0.985 & \textbf{0.996} & 0.990 \\
19 & -- & -- & -- & -- & -- & -- & 0.967 & 0.724 & 0.803 & 0.868 & 0.994 & 0.998 & 0.983 & \textbf{0.996} & 0.987 \\
20 & -- & -- & -- & -- & -- & -- & 0.966 & 0.765 & 0.815 & 0.855 & 0.996 & \textbf{0.999} & 0.979 & \textbf{0.996} & 0.985 \\
21 & -- & -- & -- & -- & -- & -- & 0.968 & 0.806 & 0.831 & 0.864 & 0.997 & \textbf{0.999} & 0.957 & 0.995 & 0.969 \\
22 & -- & -- & -- & -- & -- & -- & 0.967 & 0.850 & 0.840 & 0.788 & \textbf{0.998} & \textbf{0.999} & 0.969 & 0.982 & 0.986 \\
23 & -- & -- & -- & -- & -- & -- & 0.968 & 0.874 & 0.860 & 0.135 & 0.997 & \textbf{0.999} & 0.962 & \textbf{0.996} & 0.958 \\
24 & -- & -- & -- & -- & -- & -- & 0.967 & 0.904 & 0.869 & -- & -- & -- & -- & -- & -- \\
25 & -- & -- & -- & -- & -- & -- & 0.966 & 0.917 & 0.884 & -- & -- & -- & -- & -- & -- \\
26 & -- & -- & -- & -- & -- & -- & 0.966 & 0.939 & 0.893 & -- & -- & -- & -- & -- & -- \\
27 & -- & -- & -- & -- & -- & -- & 0.966 & 0.952 & 0.900 & -- & -- & -- & -- & -- & -- \\
28 & -- & -- & -- & -- & -- & -- & 0.965 & 0.963 & 0.909 & -- & -- & -- & -- & -- & -- \\
29 & -- & -- & -- & -- & -- & -- & 0.964 & 0.968 & 0.914 & -- & -- & -- & -- & -- & -- \\
30 & -- & -- & -- & -- & -- & -- & 0.963 & 0.974 & 0.918 & -- & -- & -- & -- & -- & -- \\
31 & -- & -- & -- & -- & -- & -- & 0.963 & 0.981 & 0.927 & -- & -- & -- & -- & -- & -- \\
32 & -- & -- & -- & -- & -- & -- & 0.963 & 0.986 & 0.931 & -- & -- & -- & -- & -- & -- \\
33 & -- & -- & -- & -- & -- & -- & 0.963 & 0.989 & 0.934 & -- & -- & -- & -- & -- & -- \\
34 & -- & -- & -- & -- & -- & -- & 0.966 & 0.992 & 0.940 & -- & -- & -- & -- & -- & -- \\
35 & -- & -- & -- & -- & -- & -- & 0.969 & 0.994 & 0.944 & -- & -- & -- & -- & -- & -- \\
36 & -- & -- & -- & -- & -- & -- & 0.972 & 0.996 & 0.948 & -- & -- & -- & -- & -- & -- \\
37 & -- & -- & -- & -- & -- & -- & 0.976 & 0.997 & 0.956 & -- & -- & -- & -- & -- & -- \\
38 & -- & -- & -- & -- & -- & -- & 0.980 & \textbf{0.998} & \textbf{0.972} & -- & -- & -- & -- & -- & -- \\
39 & -- & -- & -- & -- & -- & -- & 0.951 & 0.994 & 0.965 & -- & -- & -- & -- & -- & -- \\
\bottomrule
\end{tabular}}
\end{center}

\newpage
\section{Per-Dataset Accuracy Curves}
\label{app:accuracy_curves}
The figures below report the per-dataset accuracy curves summarized in
Figure~\ref{fig:finetune_retention}, for each pruning strategy. Dashed lines
indicate unpruned baseline accuracy and the unfilled circle at drop count 0
denotes the unpruned baseline. In Joint Layer Drop, $\bigstar$ denotes a
removed MLP sublayer and $\bullet$ denotes a removed attention sublayer.

\begin{center}
\includegraphics[width=.89\textwidth]{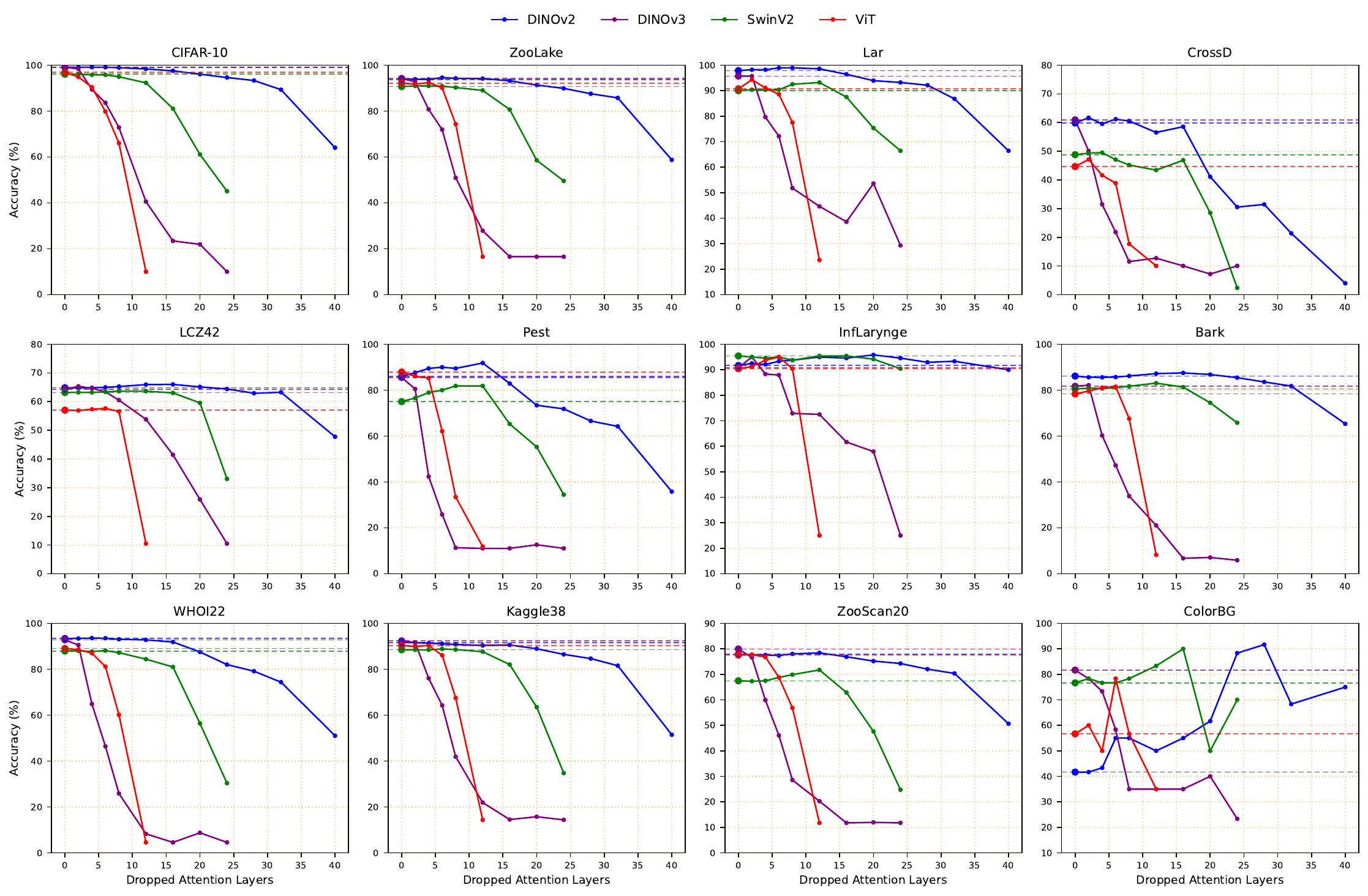}\\
{\small (a) Attention Drop}
\end{center}
\begin{center}
\includegraphics[width=.89\textwidth]{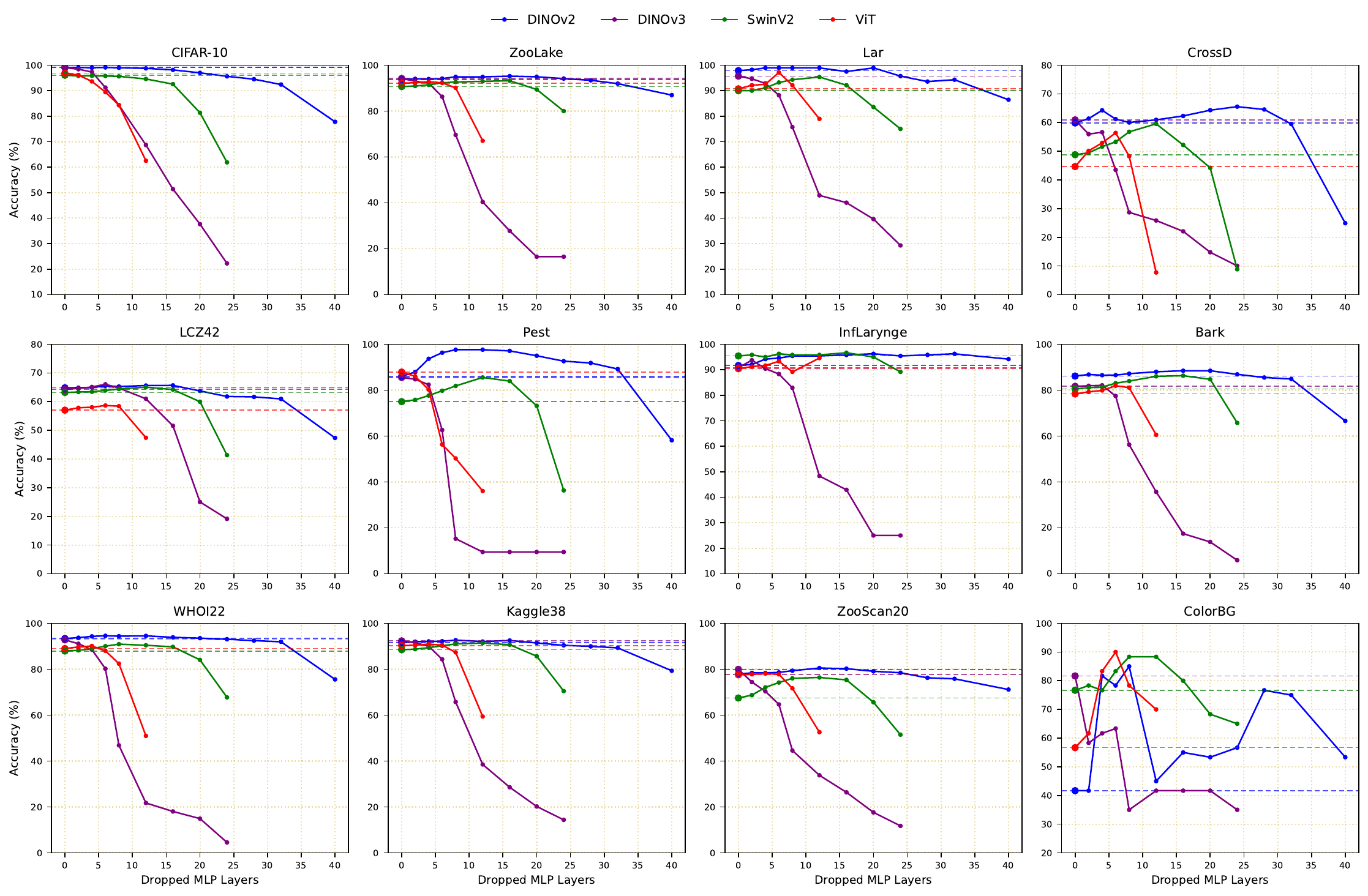}\\
{\small (b) MLP Drop}
\end{center}
\begin{center}
\includegraphics[width=.89\textwidth]{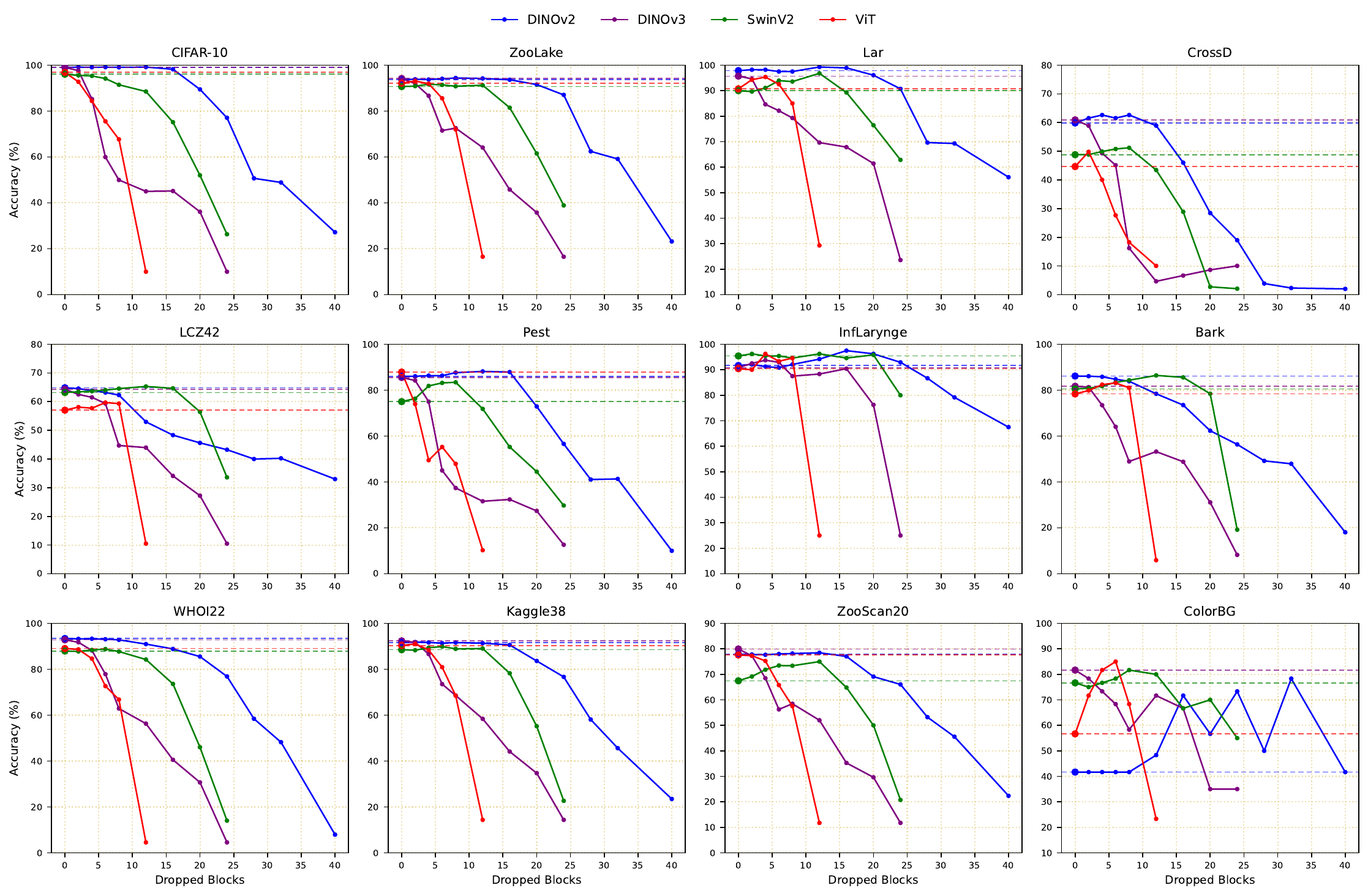}\\
{\small (c) Block Drop}
\end{center}
\begin{center}
\includegraphics[width=.89\textwidth]{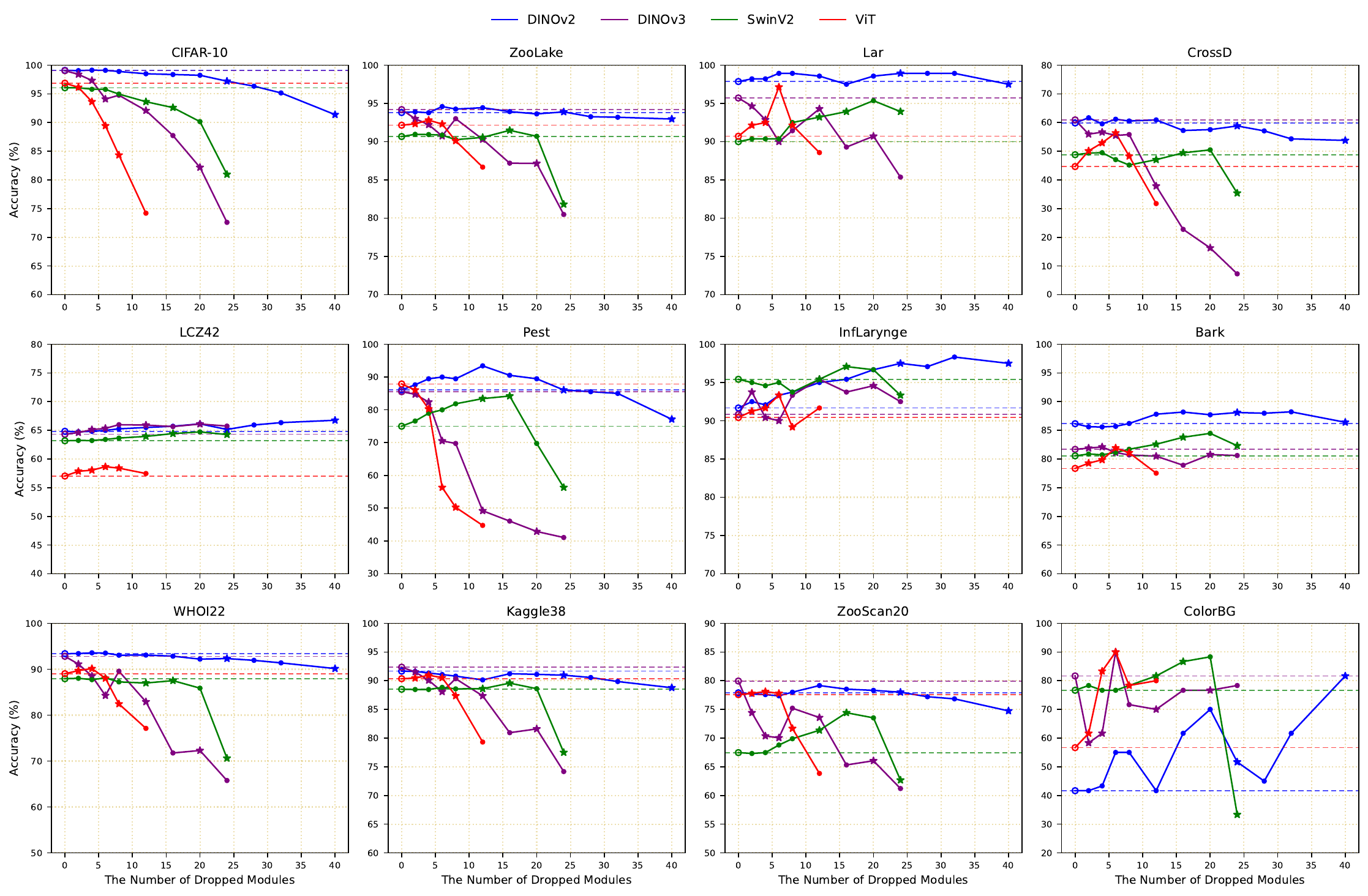}\\
{\small (d) Joint Layer Drop}
\end{center}

\newpage
\section{Macro-Averaged Metrics Across Pruning Strategies}
\label{app:radar}
The figures below report the macro-averaged precision, recall and F1-score
across the 12 downstream datasets, for each architecture and pruning
strategy, at drop counts 0, 4, 8 and 12. Each wheel corresponds to one
architecture and each spoke to one dataset, ordered by domain. Radial
gridlines mark steps of 0.1 from 0 to 1, with numeric labels omitted. A
larger enclosed area indicates better performance.

\begin{center}
\includegraphics[width=\textwidth]{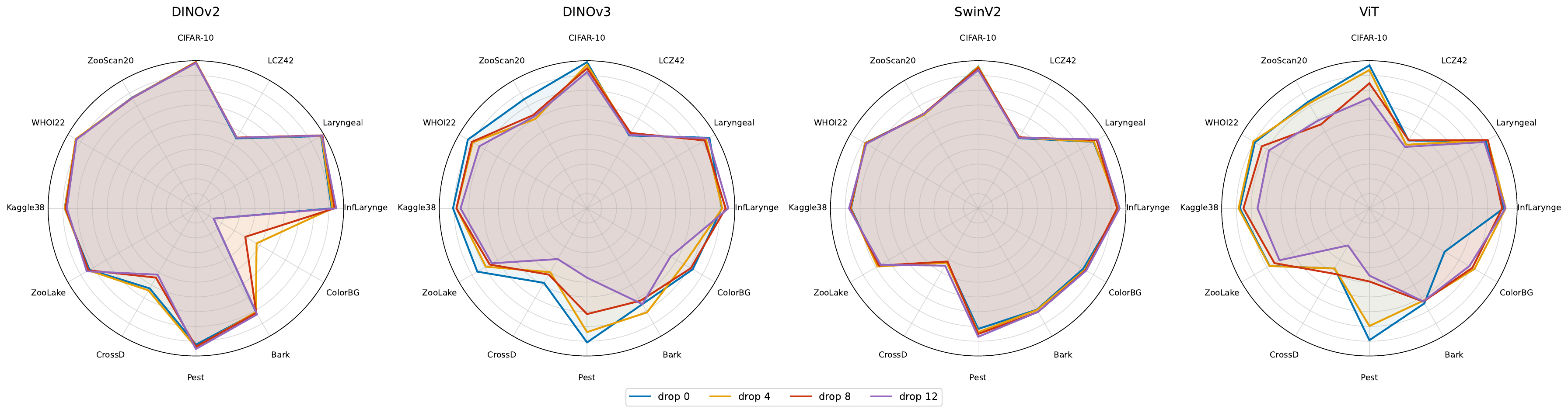}\\
{\small (a) Macro precision, Joint Layer Drop}
\end{center}
\begin{center}
\includegraphics[width=\textwidth]{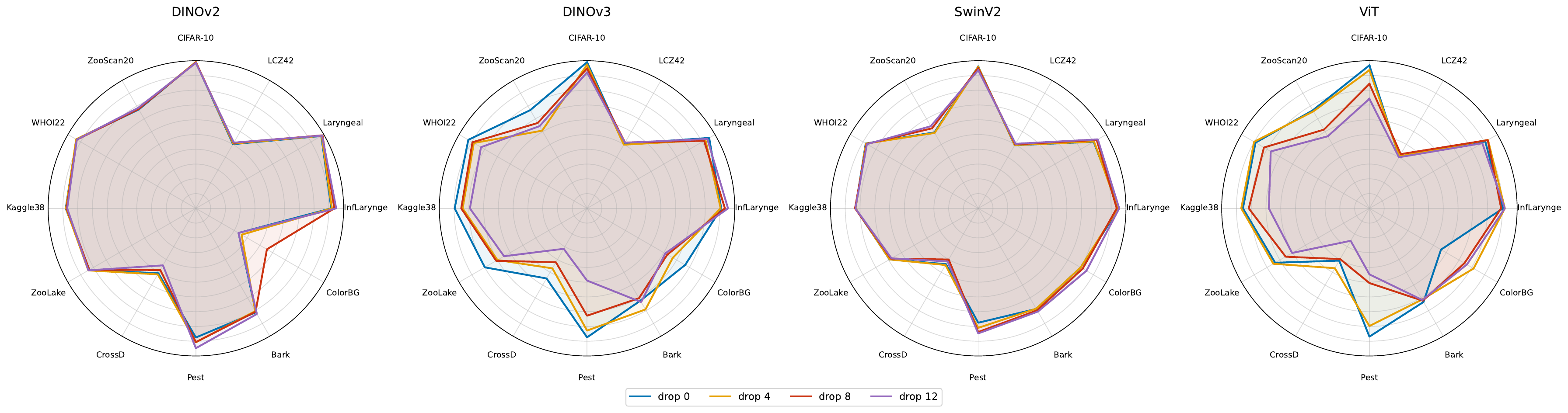}\\
{\small (b) Macro recall, Joint Layer Drop}
\end{center}
\begin{center}
\includegraphics[width=\textwidth]{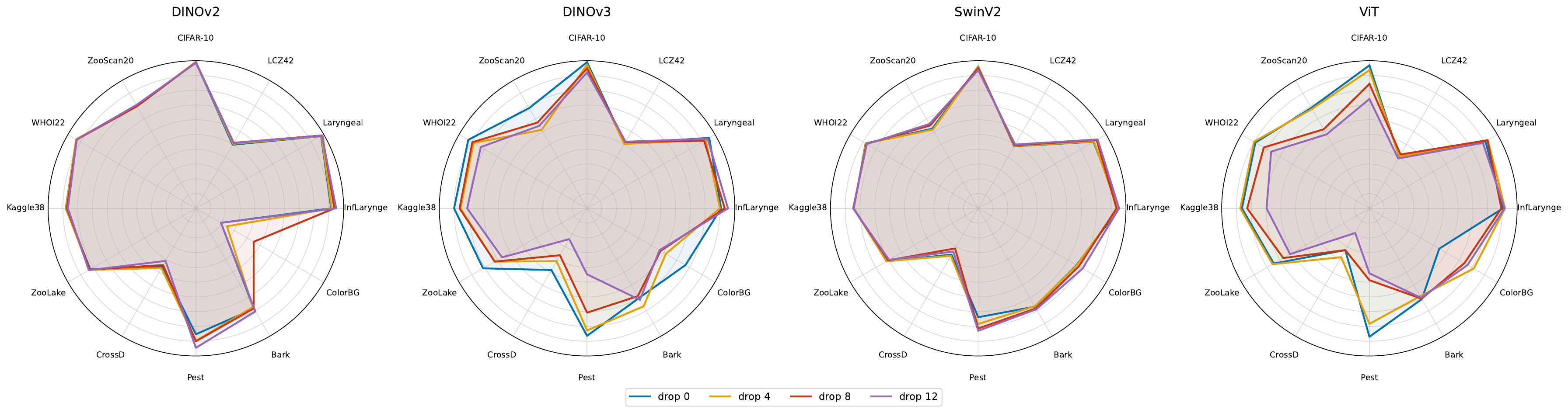}\\
{\small (c) Macro F1-score, Joint Layer Drop}
\end{center}
\begin{center}
\includegraphics[width=\textwidth]{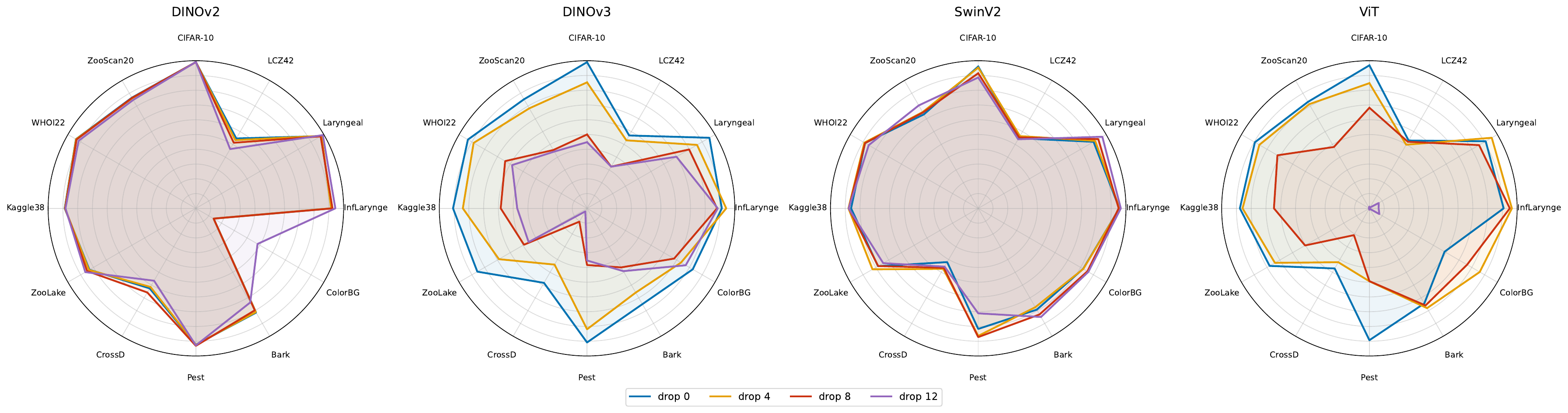}\\
{\small (d) Macro precision, Block Drop}
\end{center}
\begin{center}
\includegraphics[width=\textwidth]{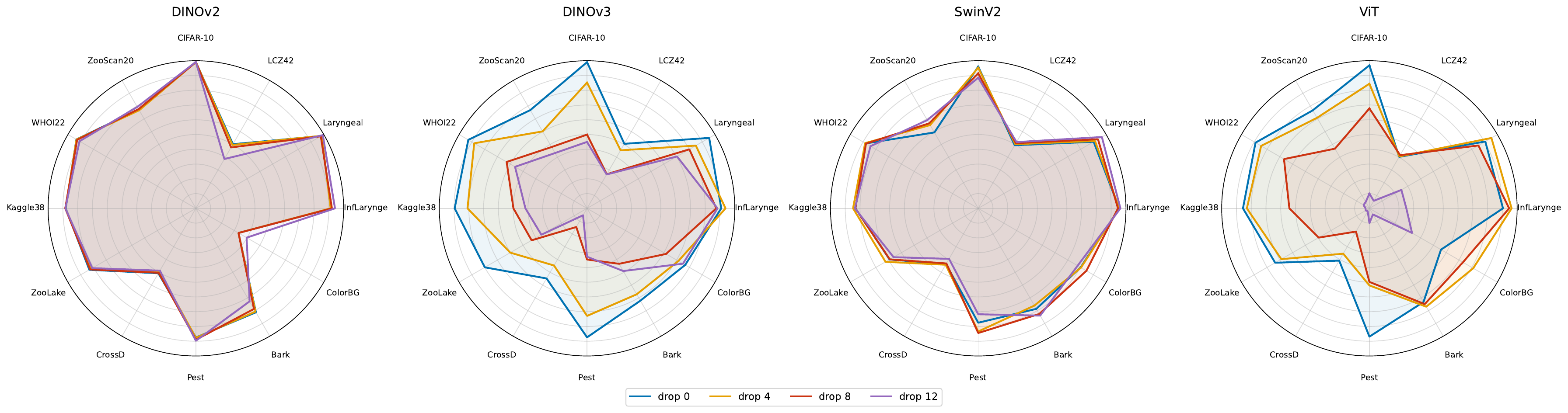}\\
{\small (e) Macro recall, Block Drop}
\end{center}
\begin{center}
\includegraphics[width=\textwidth]{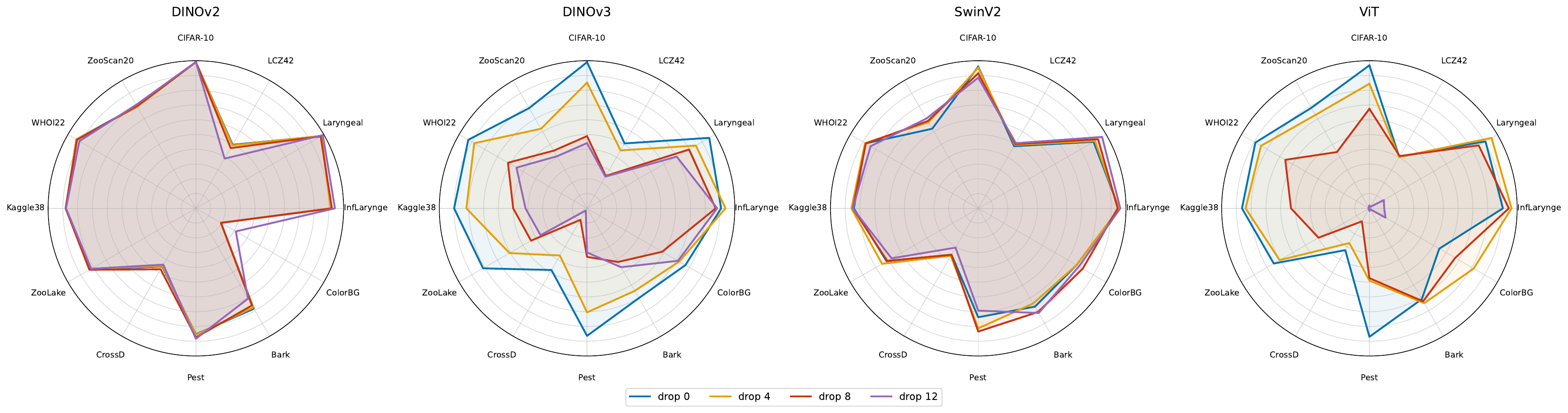}\\
{\small (f) Macro F1-score, Block Drop}
\end{center}
\begin{center}
\includegraphics[width=\textwidth]{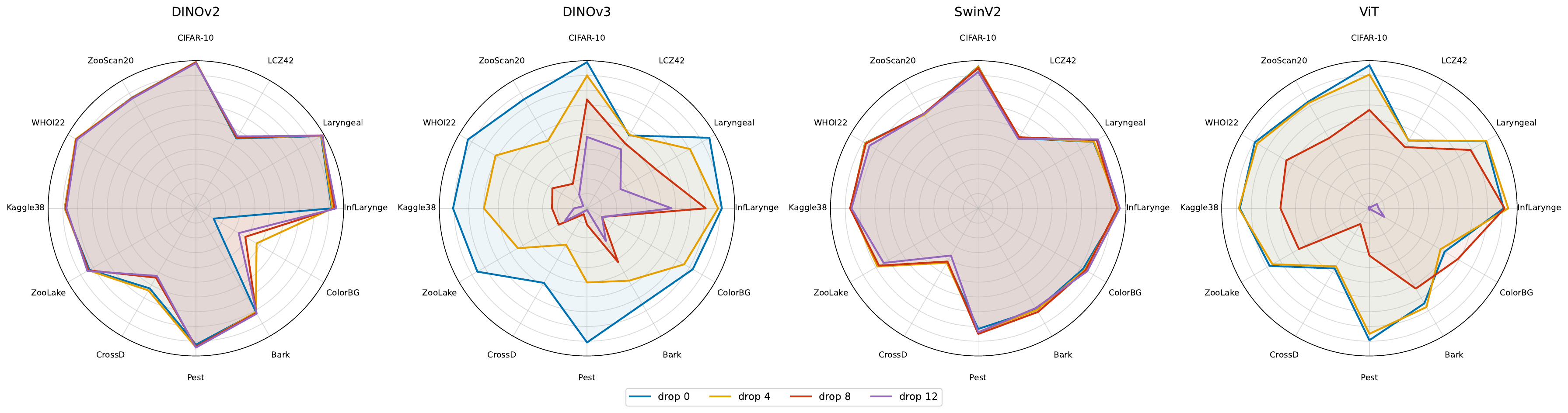}\\
{\small (g) Macro precision, Attention Drop}
\end{center}
\begin{center}
\includegraphics[width=\textwidth]{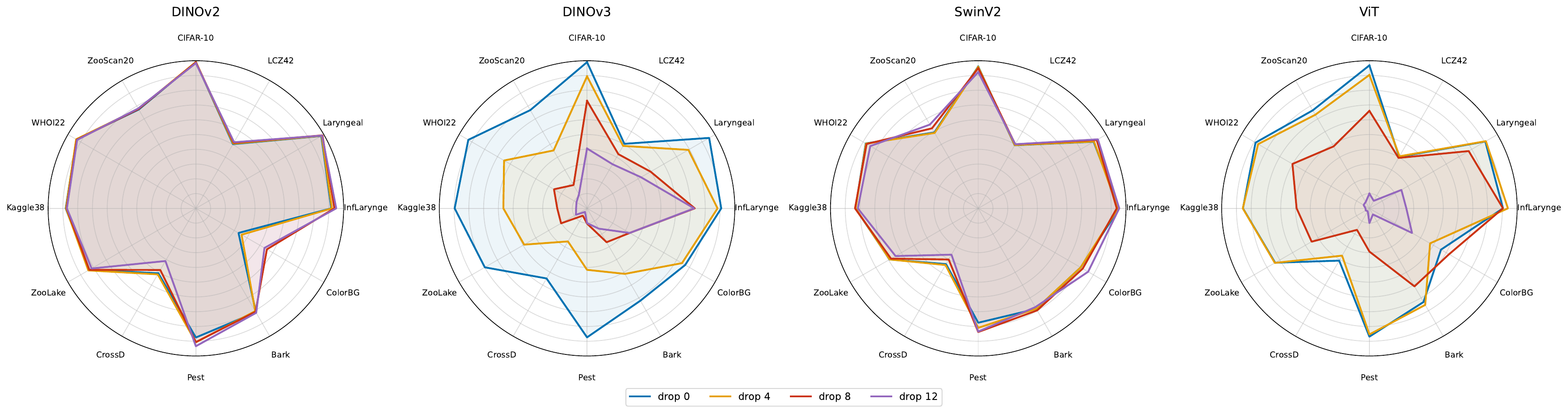}\\
{\small (h) Macro recall, Attention Drop}
\end{center}
\begin{center}
\includegraphics[width=\textwidth]{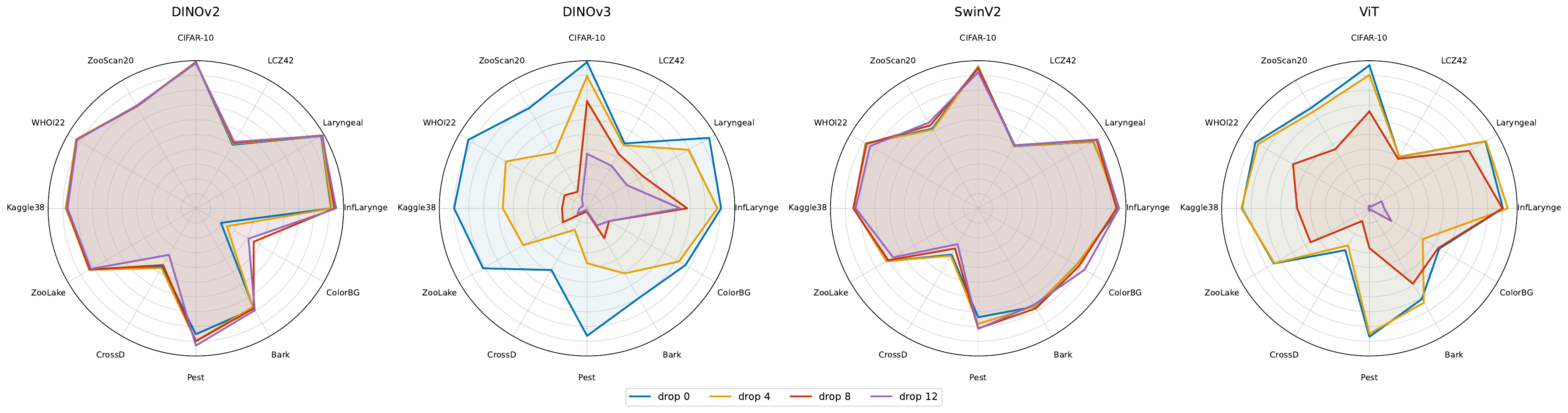}\\
{\small (i) Macro F1-score, Attention Drop}
\end{center}
\begin{center}
\includegraphics[width=\textwidth]{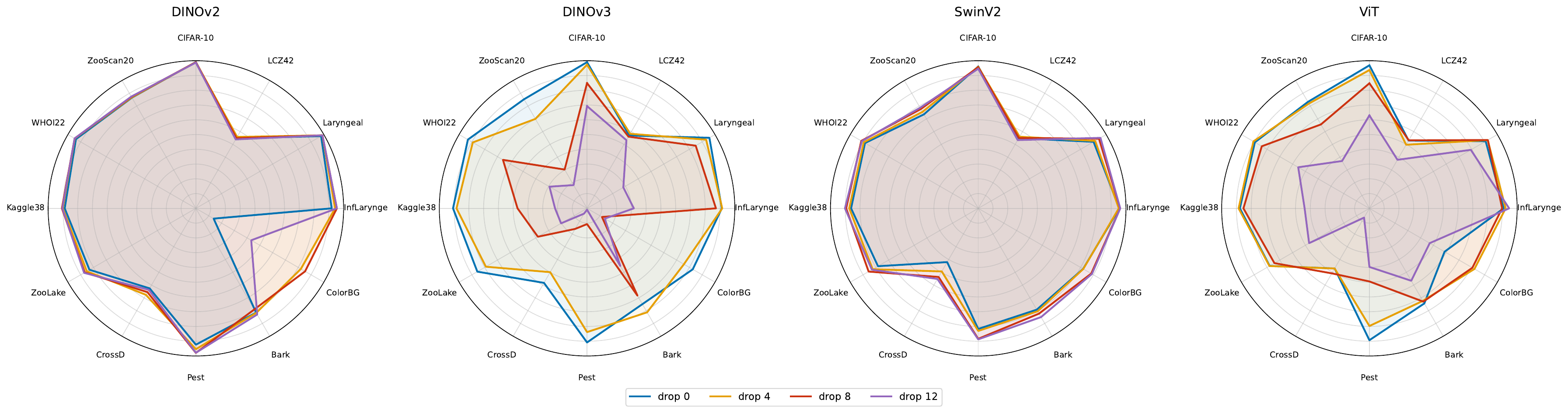}\\
{\small (j) Macro precision, MLP Drop}
\end{center}
\begin{center}
\includegraphics[width=\textwidth]{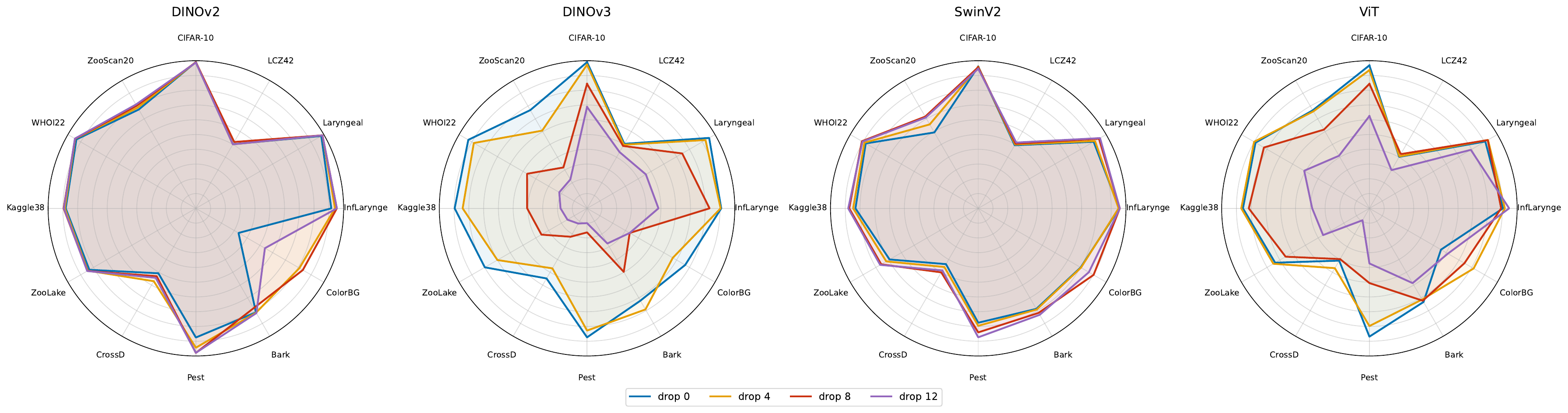}\\
{\small (k) Macro recall, MLP Drop}
\end{center}
\begin{center}
\includegraphics[width=\textwidth]{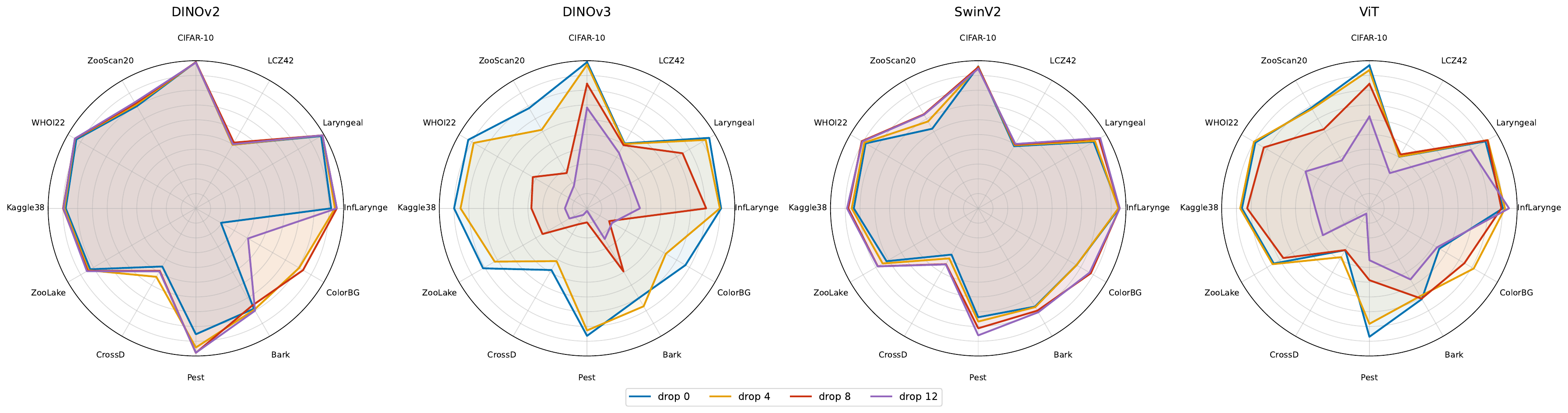}\\
{\small (l) Macro F1-score, MLP Drop}
\end{center}

\newpage
\section{Full Equivalence Analysis}
\label{app:equivalence}

\begin{center}
\captionof{table}{Difference and equivalence analysis for all 48 configurations at drop counts 4, 8 and 12 over the 12 downstream datasets. Each row reports the Holm-adjusted two-sided Wilcoxon $p$-value, the Hodges-Lehmann estimate of the median paired accuracy difference with its 90\% confidence interval, the TOST $p$-value and the resulting outcome, with the definitions of Table~\ref{tab:tost_equivalence}.}
\label{tab:tost_full}
{\scriptsize
\begin{tabular}{llc|ccc|c|l}
\toprule
\textbf{Arch} & \textbf{Method} & \textbf{Drop} & \textbf{Holm $p$} & \textbf{HL (pp)} & \textbf{90\% CI} & \textbf{TOST $p$} & \textbf{Outcome} \\
\midrule
DINOv2 & Block & 4 & 1.000 & $-0.04$ & $[-0.25, +0.17]$ & 0.000 & equivalent \\
DINOv2 & Block & 8 & 1.000 & $+0.01$ & $[-0.94, +0.61]$ & 0.001 & equivalent \\
DINOv2 & Block & 12 & 1.000 & $+0.04$ & $[-3.58, +1.43]$ & 0.117 & inconclusive \\
DINOv2 & Attn & 4 & 0.603 & $+0.05$ & $[-0.16, +0.82]$ & 0.001 & equivalent \\
DINOv2 & Attn & 8 & 0.192 & $+0.58$ & $[+0.06, +1.93]$ & 0.046 & equivalent \\
DINOv2 & Attn & 12 & 0.603 & $+0.76$ & $[-0.29, +3.16]$ & 0.170 & inconclusive \\
DINOv2 & MLP & 4 & 0.003 & $+1.38$ & $[+0.42, +4.28]$ & 0.367 & superior \\
DINOv2 & MLP & 8 & 0.003 & $+1.18$ & $[+0.71, +6.31]$ & 0.285 & superior \\
DINOv2 & MLP & 12 & 0.003 & $+1.71$ & $[+0.93, +2.80]$ & 0.311 & superior \\
DINOv2 & Joint & 4 & 0.569 & $+0.05$ & $[-0.16, +0.82]$ & 0.001 & equivalent \\
DINOv2 & Joint & 8 & 0.151 & $+0.58$ & $[+0.06, +1.93]$ & 0.046 & equivalent \\
DINOv2 & Joint & 12 & 0.151 & $+0.77$ & $[+0.06, +2.00]$ & 0.046 & equivalent \\
\midrule
DINOv3 & Block & 4 & 0.001 & $-8.19$ & $[-10.53, -5.45]$ & 0.999 & inferior \\
DINOv3 & Block & 8 & 0.001 & $-26.42$ & $[-35.27, -20.61]$ & 1.000 & inferior \\
DINOv3 & Block & 12 & 0.001 & $-31.08$ & $[-41.24, -23.20]$ & 1.000 & inferior \\
DINOv3 & Attn & 4 & 0.001 & $-16.23$ & $[-22.87, -10.54]$ & 1.000 & inferior \\
DINOv3 & Attn & 8 & 0.001 & $-46.45$ & $[-51.37, -33.65]$ & 1.000 & inferior \\
DINOv3 & Attn & 12 & 0.001 & $-56.89$ & $[-66.34, -42.34]$ & 1.000 & inferior \\
DINOv3 & MLP & 4 & 0.005 & $-2.79$ & $[-5.94, -1.35]$ & 0.788 & inferior \\
DINOv3 & MLP & 8 & 0.002 & $-27.49$ & $[-39.09, -19.65]$ & 1.000 & inferior \\
DINOv3 & MLP & 12 & 0.001 & $-46.05$ & $[-55.55, -38.18]$ & 1.000 & inferior \\
DINOv3 & Joint & 4 & 0.015 & $-2.79$ & $[-5.94, -1.35]$ & 0.788 & inferior \\
DINOv3 & Joint & 8 & 0.024 & $-3.31$ & $[-6.62, -1.30]$ & 0.883 & inferior \\
DINOv3 & Joint & 12 & 0.024 & $-6.17$ & $[-14.04, -2.54]$ & 0.968 & inferior \\
\midrule
SwinV2 & Block & 4 & 0.038 & $+0.83$ & $[+0.44, +2.62]$ & 0.102 & superior \\
SwinV2 & Block & 8 & 0.128 & $+2.03$ & $[+0.20, +4.10]$ & 0.515 & inconclusive \\
SwinV2 & Block & 12 & 0.622 & $+0.71$ & $[-2.11, +3.65]$ & 0.170 & inconclusive \\
SwinV2 & Attn & 4 & 1.000 & $+0.08$ & $[-0.11, +0.36]$ & 0.002 & equivalent \\
SwinV2 & Attn & 8 & 1.000 & $+0.44$ & $[-0.70, +1.78]$ & 0.039 & equivalent \\
SwinV2 & Attn & 12 & 1.000 & $+0.69$ & $[-1.64, +2.98]$ & 0.151 & inconclusive \\
SwinV2 & MLP & 4 & 0.010 & $+0.93$ & $[+0.34, +1.85]$ & 0.032 & equivalent \\
SwinV2 & MLP & 8 & 0.004 & $+4.04$ & $[+2.00, +6.13]$ & 0.954 & superior \\
SwinV2 & MLP & 12 & 0.004 & $+5.39$ & $[+2.39, +7.22]$ & 0.983 & superior \\
SwinV2 & Joint & 4 & 0.950 & $+0.08$ & $[-0.11, +0.36]$ & 0.002 & equivalent \\
SwinV2 & Joint & 8 & 0.950 & $+0.44$ & $[-0.70, +1.78]$ & 0.039 & equivalent \\
SwinV2 & Joint & 12 & 0.547 & $+1.21$ & $[-0.21, +3.21]$ & 0.212 & inconclusive \\
\midrule
ViT & Block & 4 & 0.791 & $-0.73$ & $[-6.34, +3.25]$ & 0.339 & inconclusive \\
ViT & Block & 8 & 0.054 & $-13.10$ & $[-23.22, -5.29]$ & 0.979 & inconclusive \\
ViT & Block & 12 & 0.001 & $-66.82$ & $[-75.80, -55.12]$ & 1.000 & inferior \\
ViT & Attn & 4 & 0.328 & $-1.15$ & $[-3.08, +0.33]$ & 0.151 & inconclusive \\
ViT & Attn & 8 & 0.010 & $-17.94$ & $[-26.96, -10.60]$ & 0.998 & inferior \\
ViT & Attn & 12 & 0.001 & $-67.46$ & $[-75.84, -53.08]$ & 1.000 & inferior \\
ViT & MLP & 4 & 0.259 & $+1.07$ & $[-0.87, +4.64]$ & 0.102 & inconclusive \\
ViT & MLP & 8 & 0.424 & $-1.98$ & $[-6.89, +2.05]$ & 0.515 & inconclusive \\
ViT & MLP & 12 & 0.015 & $-23.55$ & $[-31.81, -11.78]$ & 0.998 & inferior \\
ViT & Joint & 4 & 0.259 & $+1.07$ & $[-0.87, +4.64]$ & 0.102 & inconclusive \\
ViT & Joint & 8 & 0.424 & $-1.98$ & $[-6.89, +2.05]$ & 0.515 & inconclusive \\
ViT & Joint & 12 & 0.192 & $-7.15$ & $[-13.74, -0.83]$ & 0.935 & inconclusive \\
\bottomrule
\end{tabular}}
\end{center}

\newpage
\section{Full Random Ablation Results}
\label{app:random_ablation}
\begin{center}
\captionof{table}{Per-dataset results of the random ablation of Section~\ref{subsec:statistics}. Baseline, Similarity and Random are validation accuracies in percent after 5-epoch head fine-tuning, where Similarity removes the modules selected by the cosine-similarity ranking and Random removes the same number of modules drawn uniformly at random. Random is the mean over 3 draws with its standard deviation, and Gain is Similarity minus Random. DINOv2 Block-16 and SwinV2 Joint-8 are reported on 2 and 4 datasets, since the search phase stopped early on the others.}
\label{tab:random_ablation_full}
{\scriptsize
\begin{tabular}{lll|rrr|r}
\toprule
\textbf{Arch} & \textbf{Config} & \textbf{Dataset} & \textbf{Baseline} & \textbf{Similarity} & \textbf{Random} & \textbf{Gain} \\
\midrule
DINOv2 & Block-8 & Pest & 95.00 & 95.00 & 85.83 $\pm$ 1.18 & $+9.17$ \\
DINOv2 & Block-8 & WHOI22 & 92.73 & 92.42 & 93.33 $\pm$ 0.21 & $-0.91$ \\
DINOv2 & Block-8 & Bark & 93.55 & 92.53 & 91.85 $\pm$ 0.87 & $+0.68$ \\
DINOv2 & Block-8 & LCZ42 & 73.77 & 77.36 & 78.38 $\pm$ 0.21 & $-1.02$ \\
DINOv2 & Block-8 & CrossD & 90.62 & 90.81 & 89.46 $\pm$ 0.27 & $+1.35$ \\
DINOv2 & Block-16 & WHOI22 & 92.73 & 88.64 & 92.63 $\pm$ 0.50 & $-3.99$ \\
DINOv2 & Block-16 & CrossD & 90.62 & 87.98 & 89.05 $\pm$ 1.22 & $-1.07$ \\
DINOv2 & MLP-8 & Pest & 95.00 & 95.00 & 85.83 $\pm$ 1.18 & $+9.17$ \\
DINOv2 & MLP-8 & WHOI22 & 92.73 & 93.64 & 93.74 $\pm$ 0.56 & $-0.10$ \\
DINOv2 & MLP-8 & Bark & 93.55 & 93.58 & 92.57 $\pm$ 0.54 & $+1.00$ \\
DINOv2 & MLP-8 & LCZ42 & 78.04 & 78.17 & 78.47 $\pm$ 0.09 & $-0.30$ \\
DINOv2 & MLP-8 & CrossD & 90.62 & 92.87 & 89.28 $\pm$ 0.39 & $+3.59$ \\
DINOv2 & Joint-8 & Pest & 95.00 & 95.00 & 85.00 $\pm$ 0.00 & $+10.00$ \\
DINOv2 & Joint-8 & WHOI22 & 92.73 & 92.27 & 93.79 $\pm$ 0.33 & $-1.52$ \\
DINOv2 & Joint-8 & Bark & 93.55 & 93.66 & 91.96 $\pm$ 0.29 & $+1.69$ \\
DINOv2 & Joint-8 & LCZ42 & 78.04 & 78.67 & 78.14 $\pm$ 0.16 & $+0.53$ \\
DINOv2 & Joint-8 & CrossD & 90.62 & 90.33 & 88.69 $\pm$ 0.17 & $+1.64$ \\
DINOv2 & Joint-16 & Pest & 95.00 & 100.00 & 85.83 $\pm$ 1.18 & $+14.17$ \\
DINOv2 & Joint-16 & WHOI22 & 92.73 & 90.45 & 93.38 $\pm$ 0.29 & $-2.93$ \\
DINOv2 & Joint-16 & Bark & 93.55 & 94.01 & 91.37 $\pm$ 0.51 & $+2.63$ \\
DINOv2 & Joint-16 & LCZ42 & 78.04 & 79.48 & 78.57 $\pm$ 0.29 & $+0.91$ \\
DINOv2 & Joint-16 & CrossD & 90.62 & 90.87 & 89.64 $\pm$ 0.91 & $+1.22$ \\
\midrule
SwinV2 & Block-8 & Pest & 90.00 & 70.00 & 75.83 $\pm$ 2.36 & $-5.83$ \\
SwinV2 & Block-8 & WHOI22 & 94.24 & 87.12 & 85.40 $\pm$ 1.57 & $+1.72$ \\
SwinV2 & Block-8 & Bark & 91.37 & 91.80 & 80.56 $\pm$ 3.81 & $+11.25$ \\
SwinV2 & Block-8 & LCZ42 & 74.69 & 77.54 & 64.46 $\pm$ 3.09 & $+13.08$ \\
SwinV2 & Block-8 & CrossD & 90.46 & 89.60 & 81.47 $\pm$ 2.29 & $+8.12$ \\
SwinV2 & MLP-8 & Pest & 95.00 & 95.00 & 79.17 $\pm$ 3.12 & $+15.83$ \\
SwinV2 & MLP-8 & WHOI22 & 93.33 & 91.67 & 87.37 $\pm$ 1.04 & $+4.29$ \\
SwinV2 & MLP-8 & Bark & 90.86 & 94.19 & 84.84 $\pm$ 1.12 & $+9.35$ \\
SwinV2 & MLP-8 & LCZ42 & 73.01 & 76.15 & 70.77 $\pm$ 0.91 & $+5.38$ \\
SwinV2 & MLP-8 & CrossD & 89.63 & 91.42 & 85.13 $\pm$ 0.83 & $+6.29$ \\
SwinV2 & Joint-8 & WHOI22 & 93.33 & 91.97 & 87.02 $\pm$ 1.76 & $+4.95$ \\
SwinV2 & Joint-8 & Bark & 90.86 & 91.29 & 80.97 $\pm$ 3.15 & $+10.32$ \\
SwinV2 & Joint-8 & LCZ42 & 73.01 & 75.14 & 59.30 $\pm$ 7.58 & $+15.85$ \\
SwinV2 & Joint-8 & CrossD & 89.63 & 87.85 & 83.36 $\pm$ 0.85 & $+4.49$ \\
\midrule
ViT & Block-4 & Pest & 87.50 & 62.50 & 54.17 $\pm$ 8.25 & $+8.33$ \\
ViT & Block-4 & WHOI22 & 87.12 & 87.88 & 71.31 $\pm$ 3.31 & $+16.57$ \\
ViT & Block-4 & Bark & 83.92 & 87.42 & 74.91 $\pm$ 5.93 & $+12.51$ \\
ViT & Block-4 & LCZ42 & 53.91 & 53.24 & 50.88 $\pm$ 3.85 & $+2.35$ \\
ViT & Block-4 & CrossD & 90.42 & 90.87 & 85.13 $\pm$ 4.61 & $+5.74$ \\
ViT & MLP-4 & Pest & 87.50 & 82.50 & 60.00 $\pm$ 8.16 & $+22.50$ \\
ViT & MLP-4 & WHOI22 & 87.12 & 87.88 & 78.74 $\pm$ 4.83 & $+9.14$ \\
ViT & MLP-4 & Bark & 83.92 & 86.45 & 79.18 $\pm$ 3.98 & $+7.28$ \\
ViT & MLP-4 & LCZ42 & 53.91 & 54.42 & 50.88 $\pm$ 5.58 & $+3.53$ \\
ViT & MLP-4 & CrossD & 90.42 & 91.33 & 88.19 $\pm$ 2.61 & $+3.13$ \\
\bottomrule
\end{tabular}}
\end{center}

\newpage
\section{Full Pruning and Quantization Results}
\label{app:quantization}
This appendix reports the complete results of the stacking experiment of
Section~\ref{subsec:quantization} for all 11 configurations. The results
are split into 2 tables, since the throughput measurements are taken on
dummy inputs and therefore depend only on the model and its precision, not on
the dataset.

\begin{center}
\captionof{table}{Test accuracy and macro F1-score of the stacking experiment in FP32 for the unpruned and the pruned model of each configuration, followed by the change of each metric when the same model, with the same classification head, is cast to BF16 or quantized to 8-bit and 4-bit weights.}
\label{tab:quant_accuracy}
{\scriptsize
\setlength{\tabcolsep}{4pt}
\begin{tabular}{lll|rr|rr|rr|rr}
\toprule
 & & & \multicolumn{2}{c|}{\textbf{FP32}} & \multicolumn{2}{c|}{\textbf{$\Delta$ BF16}} & \multicolumn{2}{c|}{\textbf{$\Delta$ INT8}} & \multicolumn{2}{c}{\textbf{$\Delta$ INT4}} \\
\cmidrule(lr){4-5}\cmidrule(lr){6-7}\cmidrule(lr){8-9}\cmidrule(lr){10-11}
\textbf{Model} & \textbf{Dataset} & \textbf{Config} & Acc & F1 & Acc & F1 & Acc & F1 & Acc & F1 \\
\midrule
DINOv2 & Bark & unpruned & 90.76 & 83.3 & $-0.04$ & $-0.1$ & $+0.00$ & $-0.0$ & $-0.55$ & $-0.6$ \\
DINOv2 & Bark & Joint-40 & 90.37 & 85.5 & $+0.02$ & $+0.1$ & $+0.00$ & $+0.1$ & $+0.02$ & $+0.1$ \\
DINOv2 & InfLarynge & unpruned & 91.67 & 91.6 & $+0.00$ & $+0.0$ & $-0.42$ & $-0.4$ & $-0.42$ & $-0.4$ \\
DINOv2 & InfLarynge & Joint-40 & 97.92 & 97.9 & $+0.00$ & $+0.0$ & $+0.42$ & $+0.4$ & $+0.00$ & $+0.0$ \\
DINOv2 & WHOI22 & unpruned & 93.91 & 93.9 & $-0.09$ & $-0.1$ & $-0.06$ & $-0.1$ & $+0.12$ & $+0.1$ \\
DINOv2 & WHOI22 & MLP-8 & 94.97 & 95.0 & $-0.09$ & $-0.1$ & $-0.15$ & $-0.2$ & $-0.09$ & $-0.1$ \\
DINOv2 & CIFAR-10 & unpruned & 99.40 & 99.4 & $+0.00$ & $+0.0$ & $-0.01$ & $-0.0$ & $+0.03$ & $+0.0$ \\
DINOv2 & CIFAR-10 & Block-4 & 99.42 & 99.4 & $+0.00$ & $+0.0$ & $+0.02$ & $+0.0$ & $+0.00$ & $+0.0$ \\
DINOv2 & Laryngeal & unpruned & 98.48 & 98.5 & $+0.00$ & $+0.0$ & $+0.00$ & $+0.0$ & $+0.00$ & $+0.0$ \\
DINOv2 & Laryngeal & Joint-24 & 99.24 & 99.2 & $+0.00$ & $+0.0$ & $-0.38$ & $-0.4$ & $+0.00$ & $+0.0$ \\
DINOv2 & Zoolake & unpruned & 95.73 & 86.4 & $+0.04$ & $+0.0$ & $-0.04$ & $-0.2$ & $-0.04$ & $-0.3$ \\
DINOv2 & Zoolake & MLP-8 & 95.88 & 87.5 & $+0.04$ & $+0.2$ & $+0.00$ & $+0.0$ & $-0.04$ & $-0.1$ \\
\midrule
SwinV2 & Bark & unpruned & 82.27 & 74.7 & $+0.06$ & $+0.0$ & $+0.02$ & $+0.1$ & $-0.45$ & $-0.3$ \\
SwinV2 & Bark & MLP-12 & 86.88 & 81.9 & $-0.08$ & $-0.1$ & $+0.08$ & $+0.1$ & $-0.69$ & $-0.5$ \\
SwinV2 & Kaggle38 & unpruned & 89.74 & 86.1 & $-0.14$ & $-0.4$ & $-0.07$ & $-0.1$ & $-0.38$ & $-0.2$ \\
SwinV2 & Kaggle38 & Attn-4 & 89.98 & 86.3 & $-0.31$ & $-0.5$ & $-0.21$ & $-0.2$ & $-0.66$ & $-0.6$ \\
SwinV2 & LCZ42 & unpruned & 62.66 & 48.8 & $-1.07$ & $-1.0$ & $-1.70$ & $-1.2$ & $-10.62$ & $-11.5$ \\
SwinV2 & LCZ42 & MLP-12 & 64.71 & 50.3 & $-1.24$ & $-1.3$ & $-1.56$ & $-1.2$ & $-5.32$ & $-7.3$ \\
\midrule
ViT & Zoolake & unpruned & 91.53 & 75.2 & $+0.04$ & $+0.1$ & $-0.07$ & $-0.0$ & $-0.85$ & $-1.2$ \\
ViT & Zoolake & MLP-4 & 92.16 & 73.5 & $-0.04$ & $+0.1$ & $+0.04$ & $-0.1$ & $-1.23$ & $-0.0$ \\
\midrule
DINOv3 & LCZ42 & unpruned & 63.65 & 50.5 & $-0.49$ & $-0.5$ & $-62.80$ & $-50.4$ & $-62.55$ & $-50.3$ \\
DINOv3 & LCZ42 & Joint-20 & 65.00 & 51.3 & $-0.91$ & $-0.5$ & $-61.25$ & $-50.8$ & $-63.90$ & $-51.1$ \\
\bottomrule
\end{tabular}}
\end{center}

\newpage
\begin{center}
\captionof{table}{Weight footprint and throughput (TH) at batch size 1 on a single NVIDIA A40 of every model and precision in the stacking experiment. Reduction and Speedup are relative to the unpruned FP32 model of the same architecture. Datasets lists where each configuration was selected, and throughput does not depend on the dataset since it is measured on random inputs.}
\label{tab:quant_hardware}
{\scriptsize
\setlength{\tabcolsep}{4pt}
\begin{tabular}{lllrrrr}
\toprule
\textbf{Model} & \textbf{Config} & \textbf{Datasets} & \textbf{Weights (MB)} & \textbf{Reduction} & \textbf{TH (img/s)} & \textbf{Speedup} \\
\midrule
DINOv2 & unpruned fp32 & all 6 & 4,336 & 1.0$\times$ & 3.4 & 1.00$\times$ \\
DINOv2 & unpruned bf16 &  & 2,168 & 2.0$\times$ & 11.0 & 3.29$\times$ \\
DINOv2 & unpruned int8 &  & 1,088 & 4.0$\times$ & 4.9 & 1.47$\times$ \\
DINOv2 & unpruned int4 &  & 548 & 7.9$\times$ & 10.2 & 3.04$\times$ \\
DINOv2 & Joint-40 fp32 & Bark, InfLarynge & 2,138 & 2.0$\times$ & 6.7 & 1.98$\times$ \\
DINOv2 & Joint-40 bf16 &  & 1,069 & 4.1$\times$ & 21.6 & 6.43$\times$ \\
DINOv2 & Joint-40 int8 &  & 538 & 8.1$\times$ & 9.8 & 2.91$\times$ \\
DINOv2 & Joint-40 int4 &  & 272 & 15.9$\times$ & 20.0 & 5.96$\times$ \\
DINOv2 & Joint-24 fp32 & Laryngeal & 3,110 & 1.4$\times$ & 4.8 & 1.44$\times$ \\
DINOv2 & Joint-24 bf16 &  & 1,555 & 2.8$\times$ & 16.2 & 4.85$\times$ \\
DINOv2 & Joint-24 int8 &  & 781 & 5.6$\times$ & 7.4 & 2.21$\times$ \\
DINOv2 & Joint-24 int4 &  & 394 & 11.0$\times$ & 14.9 & 4.47$\times$ \\
DINOv2 & MLP-8 fp32 & WHOI22, Zoolake & 3,759 & 1.2$\times$ & 3.7 & 1.10$\times$ \\
DINOv2 & MLP-8 bf16 &  & 1,880 & 2.3$\times$ & 11.8 & 3.52$\times$ \\
DINOv2 & MLP-8 int8 &  & 944 & 4.6$\times$ & 5.3 & 1.58$\times$ \\
DINOv2 & MLP-8 int4 &  & 476 & 9.1$\times$ & 11.0 & 3.27$\times$ \\
DINOv2 & Block-4 fp32 & CIFAR-10 & 3,903 & 1.1$\times$ & 3.7 & 1.11$\times$ \\
DINOv2 & Block-4 bf16 &  & 1,952 & 2.2$\times$ & 12.3 & 3.67$\times$ \\
DINOv2 & Block-4 int8 &  & 980 & 4.4$\times$ & 5.6 & 1.68$\times$ \\
DINOv2 & Block-4 int4 &  & 494 & 8.8$\times$ & 11.4 & 3.40$\times$ \\
\midrule
SwinV2 & unpruned fp32 & all 3 & 343 & 1.0$\times$ & 38.1 & 1.00$\times$ \\
SwinV2 & unpruned bf16 &  & 177 & 1.9$\times$ & 38.5 & 1.01$\times$ \\
SwinV2 & unpruned int8 &  & 94 & 3.6$\times$ & 9.2 & 0.24$\times$ \\
SwinV2 & unpruned int4 &  & 53 & 6.5$\times$ & 16.7 & 0.44$\times$ \\
SwinV2 & MLP-12 fp32 & Bark, LCZ42 & 223 & 1.5$\times$ & 40.0 & 1.05$\times$ \\
SwinV2 & MLP-12 bf16 &  & 117 & 2.9$\times$ & 39.5 & 1.04$\times$ \\
SwinV2 & MLP-12 int8 &  & 64 & 5.3$\times$ & 10.5 & 0.28$\times$ \\
SwinV2 & MLP-12 int4 &  & 38 & 9.1$\times$ & 18.3 & 0.48$\times$ \\
SwinV2 & Attn-4 fp32 & Kaggle38 & 313 & 1.1$\times$ & 43.3 & 1.10$\times$ \\
SwinV2 & Attn-4 bf16 &  & 161 & 2.1$\times$ & 43.1 & 1.09$\times$ \\
SwinV2 & Attn-4 int8 &  & 86 & 4.0$\times$ & 10.6 & 0.27$\times$ \\
SwinV2 & Attn-4 int4 &  & 48 & 7.2$\times$ & 18.9 & 0.48$\times$ \\
\midrule
ViT & unpruned fp32 & Zoolake & 327 & 1.0$\times$ & 206.3 & 1.00$\times$ \\
ViT & unpruned bf16 &  & 164 & 2.0$\times$ & 204.0 & 0.99$\times$ \\
ViT & unpruned int8 &  & 83 & 4.0$\times$ & 32.7 & 0.16$\times$ \\
ViT & unpruned int4 &  & 42 & 7.8$\times$ & 58.7 & 0.28$\times$ \\
ViT & MLP-4 fp32 & Zoolake & 255 & 1.3$\times$ & 222.9 & 1.08$\times$ \\
ViT & MLP-4 bf16 &  & 128 & 2.6$\times$ & 222.1 & 1.08$\times$ \\
ViT & MLP-4 int8 &  & 65 & 5.1$\times$ & 36.4 & 0.18$\times$ \\
ViT & MLP-4 int4 &  & 33 & 9.9$\times$ & 65.7 & 0.32$\times$ \\
\midrule
DINOv3 & unpruned fp32 & LCZ42 & 1,156 & 1.0$\times$ & 64.2 & 1.00$\times$ \\
DINOv3 & unpruned bf16 &  & 578 & 2.0$\times$ & 71.9 & 1.12$\times$ \\
DINOv3 & unpruned int8 &  & 290 & 4.0$\times$ & 12.3 & 0.19$\times$ \\
DINOv3 & unpruned int4 &  & 146 & 7.9$\times$ & 26.2 & 0.41$\times$ \\
DINOv3 & Joint-20 fp32 & LCZ42 & 660 & 1.8$\times$ & 107.5 & 1.67$\times$ \\
DINOv3 & Joint-20 bf16 &  & 330 & 3.5$\times$ & 113.1 & 1.76$\times$ \\
DINOv3 & Joint-20 int8 &  & 166 & 7.0$\times$ & 20.0 & 0.31$\times$ \\
DINOv3 & Joint-20 int4 &  & 84 & 13.8$\times$ & 43.0 & 0.67$\times$ \\
\bottomrule
\end{tabular}}
\end{center}

\end{document}